\documentclass[lettersize,journal]{IEEEtran}
\usepackage{amsmath,amsfonts}
\usepackage{algorithmic}
\usepackage{algorithm}
\usepackage{array}
\usepackage{textcomp}
\usepackage{stfloats}
\usepackage{url}
\usepackage{verbatim}
\usepackage{graphicx}
\usepackage{cite}
\usepackage{amssymb}
\usepackage{float}
\usepackage{subfigure}
\usepackage{diagbox}
\usepackage{wrapfig}
\usepackage{booktabs}
\usepackage{multirow}
\usepackage{color}
\usepackage{arydshln}

\begin{document}

\title{Zero-shot Class Unlearning via Layer-wise Relevance Analysis and Neuronal Path Perturbation}

\author{Wenhan~Chang,
        Tianqing~Zhu,~\IEEEmembership{Member,~IEEE},
        Ping Xiong*,
        Yufeng Wu,
        Faqian Guan,
        Wanlei Zhou,~\IEEEmembership{Life Fellow,~IEEE}
\IEEEcompsocitemizethanks{
    \IEEEcompsocthanksitem Wenhan Chang and Ping Xiong are with the School of Information Engineering, Zhongnan University of Economics and Law. 
    \IEEEcompsocthanksitem Yufeng Wu is with the China University of Geosciences, Wuhan, 
    \IEEEcompsocthanksitem Tianqing Zhu, Faqian Guan and Wanlei Zhou are with the City University of Macau.
    \IEEEcompsocthanksitem Tianqing Zhu is the corresponding author. E-mail: tqzhu@cityu.edu.mo}
}

\markboth{Journal of \LaTeX\ Class Files,~Vol.~14, No.~8, August~2021}%
{Shell \MakeLowercase{\textit{et al.}}: A Sample Article Using IEEEtran.cls for IEEE Journals}


\maketitle

\begin{abstract}

Machine unlearning is a technique that removes specific data influences from trained models without the need for extensive retraining. However, it faces several key challenges, including the lack of explanation, privacy concerns during the unlearning process, and the high demand for time and computational resources. This paper presents a novel unlearning approach to tackle above challenges by employing Layer-wise Relevance Analysis and Neuronal Path Perturbation. Our method balances machine unlearning performance and model utility by identifying and perturbing highly relevant neurons, thus achieving effective unlearning. Using unseen data that has not been presented in the original training set, our method achieves zero-shot unlearning, which allows for the removal of specific class knowledge without accessing the original training data during the unlearning process. This approach ensures robust privacy protection. Experimental results demonstrate that our approach effectively removes targeted data from the target unlearning model while maintaining the model's utility, offering a practical solution for privacy-preserving machine learning.

\end{abstract}


\section{Introduction}
\label{sec:introduction}

\IEEEPARstart{I}{n} recent years, with the growing public attention on Artificial Intelligence (AI) models, research has increasingly focused on privacy concerns stemming from user-model interactions. Once a model is trained, users may wish to withdraw their data and remove its influence from the model. Traditionally, this can be achieved by retraining the model, but retraining is time consuming and inefficient for large datasets, such as those used in large language models (LLM)~\cite{chang2024classmachineunlearningcomplex}. The goal of \textbf{unlearning} is to address this issue by selectively removing the model’s memory of specific data without retraining the entire model. This can be done through data reorganization or model manipulation, allowing compliance with data deletion requests and safeguarding user privacy \cite{10.1145/3603620, chang2024zeroshotclassunlearninglayerwise}.

Machine unlearning research is primarily divided into class unlearning and sample unlearning~\cite{10.1145/3603620,nguyen2022survey}. Class unlearning aims to eliminate all knowledge the model has acquired from a specific data class. This type of unlearning is often necessary when data usage policies require the deletion of certain users' data, when the data has lost its relevance, or when it has introduced bias. 
In contrast, sample unlearning focuses on removing the influence of specific data samples from the model. This approach is commonly applied when users request the deletion of their data or when problematic data samples need to be excluded. 

Class unlearning methods have been applied to a wide range of AI tasks. For example, Chen et al.~\cite{chen2023boundary} achieved unlearning in image classification models by manipulating the model's decision space. Ma et al.~\cite{ma2022learn} implemented unlearning using neuron masks derived from gradients. Furthermore, Sun et al.~\cite{sun2023generativeadversarialnetworksunlearning} employed unlearning methods to make StyleGAN forget specific knowledge classes. Additionally, Chen et al.~\cite{chen2022graph} proposed an unlearning framework for Graph Neural Networks (GNNs). These examples illustrate that class unlearning goes beyond simply removing individual samples—it ensures that entire problematic or sensitive data categories can be eliminated, which is essential for building fair and privacy-preserving AI systems.


As we explore the common issue of class unlearning in neural networks, a fundamental question arises: Can we achieve knowledge unlearning without compromising model utility? Delving into the challenges associated with balancing unlearning and model utility reveals three specific challenges:

\begin{itemize}
    \item \textbf{Lack of explainable machine unlearning methods.} A prominent challenge in machine unlearning is the need for more basic explanations regarding the unlearning principle. Even though users submit unlearning requests to model owners, there's still a significant lack of clear explanations on how the unlearning methods influence the models' parameters. 
    
    \item \textbf{Model owners may need to access users' original data to use unlearning methods.} Many existing methods use the users' original data for unlearning, either by fine-tuning or by directly modifying the parameters. Using original data raises serious privacy concerns because it may expose sensitive information during model updates.
    
    \item \textbf{High cost on time and computational resources} Many existing unlearning methods necessitate extensive calculations, which can involve retraining or fine-tuning models based on large datasets. 
\end{itemize}

We address the three challenges by investigating the relevance between the model's neurons and the unlearning information.``Relevance" refers to the contribution or importance of each neuron in the neural network to the final prediction. We identify the path of neurons with high relevance, which we define as the classification path. Our goal is to determine the neuron classification path by assessing the relevance of each neuron to the unlearning information. Through layer-wise relevance analysis, we can explain the internal principles of machine unlearning.

For the second challenge, our solution follows the concept of \textit{zero-shot machine unlearning} as defined by Chundawat et al.~\cite{10097553}. In their work, zero-shot unlearning refers to performing the unlearning process without needing access to the original training dataset (including retain data and forget data), rather than not using any data. Based on this definition, our method utilizes ``unseen samples", which are data not present in the original training dataset, to perform the unlearning process. These unseen samples can be either directly extracted from the test set or generated by generative models. By relying solely on such data, we effectively adhere to the zero-shot unlearning paradigm and help ensure the privacy of users' original training data during the machine unlearning process.

In terms of the method's implementation efficiency, the proposed method does not need to modify the overall model parameters, significantly mitigate the time and computational cost. We analyze the data related to the unlearning class information and identify the neurons that show high relevance. Based on this, our method applies Neuronal Path Perturbation by selectively disconnecting these neurons within a single fully connected layer from their connections to preceding and succeeding layers. This action perturbs the model's classification path, causing it to lose the ability to classify the unlearning class data. Due to the inherent redundancy in neural network paths for other classes and our perturbation being limited to a single layer, the model's overall utility is largely preserved. 

In summary, we introduce layer-wise relevance analysis and neuronal path perturbation to fix the privacy and effectiveness challenges. To ensure robust privacy protection, we minimize the exposure of model owners to users' data. Also, we implement unlearning techniques within the neural network architecture to balance unlearning effectiveness and model utility. The key contributions of this paper can be summarized as follows:

\begin{itemize}
    \item We propose an explainable zero-shot unlearning method significantly improving unlearning speed and reducing resource consumption. Using layer-wise relevance analysis and neuronal path perturbation, we can adjust neuron weights highly related to unlearning information within the model. Thereby, we can achieve effective unlearning without compromising the model’s utility.
    
    \item Our method can achieve class unlearning without accessing the user's original data. We extract data closely related to the unlearning information from AI-generated contexts or datasets without samples from the training set to identify neurons within the neural network that are relevant to unlearning information without users' data.
    
    \item We demonstrate that unlearning in neural networks can be approached from the perspective of network inference by conducting experiments. Our method addresses traditional parameter adjustments and delves into the redundant characteristics within neural networks. 
\end{itemize}

\section{Related Works}

\subsection{Machine Unlearning}

As the importance of data privacy protection continues to rise, class unlearning has emerged as a critical technique in machine unlearning, gaining increasing attention from researchers. However, achieving this goal is not straightforward; thoroughly exploring existing unlearning methods can provide valuable insights for our research and enhance our understanding of the challenges.

First, Bourtoule et al.~\cite{9519428} introduced the SISA training framework, which accelerates unlearning by limiting the impact of data points during training, making it easier for users to revoke data access and request deletion after sharing.
Chen et al.~\cite{chen2023boundary} developed Boundary Unlearning, which removes a class from a DNN by shifting its decision boundary, enabling rapid unlearning. 
Jia et al.~\cite{10.5555/3666122.3668368} proposed a model-based approach to unlearning, using weight pruning for model sparsification in a "prune first, then unlearn" method. 
Researchers can also use unlearning methods to remove model bias. 
Chen et al.~\cite{10.5555/3666122.3666761} introduced the FMD framework, which identifies and removes biases from trained models using counterfactual concepts and influence functions. 
Kurmanji et al.~\cite{10.5555/3666122.3666217} proposed SCRUB, a versatile unlearning algorithm for removing biases, resolving confusion, and protecting user privacy across various applications. 
Foster et al.~\cite{Foster_Schoepf_Brintrup_2024} introduced Selective Synaptic Dampening (SSD), a retrain-free unlearning method that uses the Fisher information matrix to dampen parameters critical to forgotten data. 
Liu et al.~\cite{Liu_Wang_Huai_Miao_2024} explored security risks in unlearning, proposing two backdoor attack methods: one exploiting unlearning requests without data poisoning, and another involving poisoned data with a trigger activated by malicious unlearning. 

However, most unlearning methods require the use of original training data, which increases the frequency of privacy data exposure and the risk of leakage. As a result, researchers have begun to focus on unlearning methods that do not rely on the training data. For example, Chundawat et al~\cite{10097553} introduced zero-shot machine unlearning, tackling unlearning without access to original training data. They propose two methods—error-minimizing noise and gated knowledge transfer—to remove the influence of forgotten data while preserving model performance on retained data. 
Tarun et al.~\cite{10113700} presented a fast, scalable unlearning framework using error-maximizing noise and weight manipulation. Their impair-repair approach unlearns specific data classes without the full training dataset, followed by a repair step to recover model performance. 
Existing research methods have also extended to feature unlearning. 

Beyond the class or sample level, machine unlearning research is expanding. Researchers are now also conducting more thorough investigations at the feature level, seeking more detailed privacy protection. Xu et al.~\cite{10607903} introduced a ``feature unlearning" scheme to address the limitations of existing machine unlearning methods, which typically focus on removing entire instances or classes. Recognizing the need for feature-level unlearning to maintain model utility. They further proposed an approach~\cite{xu2024efficienttargetlevelmachineunlearning} constructed an essential graph to describe relationships between critical parameters, identify parameters important to both the forgotten and retained targets and use a pruning-based technique to remove information about the target to be unlearned effectively. 

Our approach differs significantly by operating at the neuron level, focusing on the relevance of individual neurons to the target class to be unlearned. This enables a highly fine-grained and precise unlearning process, thereby enhancing its overall granularity. Furthermore, our method prioritizes user privacy by not requiring access to the original training data, effectively achieving zero-shot unlearning. This distinct approach ensures efficient unlearning while effectively maintaining model utility.

\subsection{Layer-wise Relevance Propagation}

Layer-wise Relevance Propagation (LRP) aims to provide interpretability for deep learning models by attributing model predictions back to input features. It redistributes the output relevance scores to highlight which features are most influential in driving the model's decisions, enhancing transparency and trust in its behavior. 

Bach et al.~\cite{10.1371/journal.pone.0130140} contributed to the development of layer-wise relevance propagation, a methodology that enables pixel-wise decomposition of classification decisions, enhancing the interpretability of automated image classification systems. 
Binder et al.~\cite{10.1007/978-3-319-44781-0_8} proposed an approach to extend layer-wise relevance propagation to neural networks with local renormalization layers, which is a very common product-type non-linearity in convolutional neural networks. 

Today, research on LRP still holds a significant influence. Li et al.~\cite{LI2025110956} proposed the Weight-dependent Baseline Layer-wise Relevance Propagation algorithm, which improves instance-level explanations by addressing limitations in existing LRP methods that disregard model weights or sample features.  Hatefi et al.~\cite{hatefi2024pruningexplainingrevisitedoptimizing} proposed optimizing the hyperparameters of attribution methods from eXplainable AI for network pruning, demonstrating improved model compression rates for both transformer-based and convolutional networks.

Based on the above work, Layer-wise Relevance Propagation can quantitatively analyze the relevance between neurons in a neural network and specific data categories. This helps us identify which neurons are associated with the categories of unlearning data, allowing for more precise and efficient completion of the unlearning process.

\section{Backgrounds}

\begin{table}[ht]
\centering
\label{tab:my_label}
\caption{Abbreviations and Acronyms. This table provides explanations for the abbreviations and acronyms used in the article.}
\begin{tabular}{cl}
\toprule
\textbf{Symbol} & \textbf{Description} \\
\midrule
$F$ & Model’s prediction function \\
$L$ & Accuracy loss function \\
$\theta$ & Model's parameters \\
$\theta^{\prime}$ & Approximate parameters \\
$D/D'$ & Training/Updated dataset \\
$N_c$ & Number of classes \\
$C_i$ & Class $i$ \\
$(x,y)$ & Input-output pairs \\
$\eta$ & Learning rate \\
$\nabla$ & Gradient \\
$r$ & Propagation rules \\
$w$ & Weight \\
$p$ & Neuron position \\
$R$ & Relevance score \\
$l$ & Layer $l$ \\
$m_i$ & Amount of neurons in Layer $i$ \\
$m_p$ & Amount of perturbed neurons in Layer $i$ \\
$G$ & Gigabyte \\
\bottomrule
\end{tabular}
\end{table}

\subsection{Machine Unlearning}

Machine unlearning is the process of removing specific data's influence from a trained machine-learning model. 

Consider a machine learning model with a prediction function $F$, trained on a dataset $D$ via optimizing an accuracy loss function $L$. The model's parameters after training are represented by $\theta$. When we need to unlearn specific data $(x_u,y_u)$, we update the training dataset to $D'$ by removing the data that needs to be forgotten as Equation~\ref{unlearning_dataset}. 

\begin{equation}
    D'=D\setminus\{(x_u,y_u)\}
\label{unlearning_dataset}
\end{equation}

As shown in Equation~\ref{retrain}, the updated dataset $D'$ is then used to retrain the model, resulting in new parameters $\theta'$ that ideally do not retain any influence from the removed data.

\begin{equation}
    \theta^{\prime}\approx\operatorname{argmin}_\theta\sum_{(x,y)\in D'}L(F(x;\theta),y)
\label{retrain}
\end{equation}

\subsection{Layer-wise Relevance Propagation}

Layer-wise Relevance Propagation is a technique used to interpret the decisions made by deep neural networks. Let $F(x)$ represent the model’s prediction function for an input $x$. Given a neural network, LRP aims to attribute the prediction $F(x)$ to the individual input features by propagating relevance scores through the network layers.

For a given class $C_i$, the relevance score $R^o$ at the output layer $l_o$ is initialized as the prediction for that class. If the neural network outputs a probability distribution over classes, the relevance score $R_j^{o}$ for each output neuron $j$ corresponding to class $C_i$ can be initialized as:

\begin{equation}
    R_j^o=\delta_{ji}\cdot F_{C_i}(x),\ \delta_{ij} =
        \begin{cases} 
        1 & \text{if } j = i \\
        0 & \text{otherwise}
        \end{cases}
\label{relevance_initiation}
\end{equation}

In Equation~\ref{relevance_initiation}, $\delta$ is the Kronecker delta function, which is $1$ if $j=i$ and $0$ otherwise, and $F_{C_i}(x)$ is the output of the network for class $C_i$. This initialization means the relevance score is set to the model's output for the target class and zero for other classes.

The relevance scores are propagated from the output layer back to the input layer using a set of propagation rules. Let $R_k^{l}$ be the relevance score of neuron $k$ in layer $l$, and we denote all the different neurons in layer $l$ as $k'$. The relevance score is propagated from layer $l+1$ to the layer $l$ according to the following general rule:

\begin{equation}
R_k^l=\sum_j\frac{z_{kj}}{\sum_{k^{\prime}}z_{k^{\prime}j}}R_j^{l+1},\ k\in l,\ k'\in l,\ j\in l+1
\end{equation}

where $z_{kj}=a_k\cdot w_{kj}$ represents the contribution of neuron $k$ to $j$, with $a_k$ being the activation of neuron $k$ and $w_{kj}$ being the weight connecting neuron $k$ to $j$.

\section{Methodology}
\label{sec:method}

\subsection{Threat Model}

\textbf{Purpose}: Our method aims to achieve class unlearning efficiently. We strive to minimize disruption to the model's utility during unlearning. Overall, our method seeks to balance the unlearning results with the utility of the unlearned model.

\textbf{Users}: Users can know that their data is used to train the model and might be aware of privacy policies and data usage agreements. They can request the unlearning of that specific class and withdraw consent for its use in the model. After requesting class unlearning, users may receive confirmation that the model's knowledge of data from that class has been removed.

\textbf{Model owner}: Model owners cannot directly access user data and only have insight into the class distribution of the training dataset. They understand the model's performance metrics and how unlearning affects it. They can locate and delete specific user data upon request, retrain, or adjust the model to maintain its performance, and ensure compliance with relevant laws. They may need to provide users with proof of unlearning.

\subsection{Overview of Our Method}

\begin{figure*}[htbp]
\centering
\includegraphics[width=1.0\linewidth]{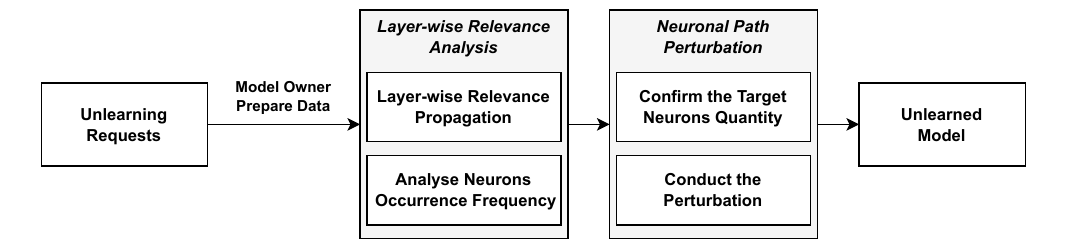} 
\caption{The overview of the unlearning method. The flowchart outlines the process from data preparation and layer-wise relevance analysis to neuronal path perturbation. Specific steps include preparing data, using layer-wise relevance propagation to analyze neuron occurrence frequency, confirming the target number of neurons, and conducting neuronal path perturbation to create an unlearned model.}
\label{overview}
\end{figure*}





Our method for achieving machine unlearning comprises several steps. As illustrated in Figure~\ref{overview}, after receiving unlearning requests from users, the model owner prepares dataset $D'$, extracted from the test set, for subsequent processing. The two primary steps involved are Layer-wise Relevance Analysis and Neuronal Path Perturbation.

\textbf{Layer-wise Relevance Analysis}.
This step obtains and analyzes the relevance of all neurons to the classification task using Layer-wise Relevance Propagation. According to LRP rules $r$, the contributions of the model's inputs to the predictions are determined through backward propagation through the network layers. By comparing the relevance scores $R$ of neurons in each layer, we can assess their importance to the classification. As depicted in Figure~\ref{overview}, analyzing the relevance for each unlearning sample allows us to identify neurons that are frequently highly relevant or critical for the target classification, especially considering the specific characteristics of the samples to be unlearned.

\textbf{Neuronal Path Perturbation}.
This step involves perturbing selected neurons, particularly those identified as relevant or critical, to achieve unlearning. Perturbation can be applied, for example, to neurons in the fully connected layer preceding the output layer. Perturbing only the output layer would be akin to simply suppressing the model's output for a specific class, rather than truly unlearning. As illustrated in Figure~\ref{overview}, we first determine the quantity $m_p$ of target neurons to be perturbed. We then perturb these neurons to disrupt or sever the model's neuronal pathways responsible for the target classification, thereby achieving machine unlearning.

Following these two steps, a class-unlearned model is obtained without requiring additional training. The details of our method are described in the following section.

\subsection{Layer-wise Relevance Analysis}

A neural network can be viewed as a complex function that performs a specific task. Network predictions result from internal interactions among its neurons. To achieve unlearning without additional training, we need to understand the internal workings of neural networks. However, obtaining the relevance of neurons to the model's inputs is challenging due to the black-box nature of neural networks.

To address this challenge, we first apply LRP to obtain an internal relevance score $R$ for each neuron. We utilize dataset $D'$ to conduct the relevance analysis. While LRP is typically applied to inputs, here we use $D'$ (which contains data related to the unlearning class) to analyze relevance patterns across multiple relevant samples. The relevance $R$ is conserved through the layers during backpropagation, ensuring that the total relevance is preserved from the output layer back to the input layer. This backward propagation process can be described by Equation~\ref{lrp_process_i}:

\begin{equation}
    R_i^{l}=\sum_{j\in(l+1)}R_{i\leftarrow j}^{l,l+1}
\label{lrp_process_i}
\end{equation}
Here, based on LRP, $R_i^{l}$ is the relevance assigned to neuron $i$ at layer $l$, and $R_{i\leftarrow j}^{l,l+1}$ is the portion of relevance propagated from neuron $j$ at layer $l+1$ to neuron $i$ at layer $l$. For layer $l$, the total relevance of neuron $i$ ($R_i^l$) is computed by summing all relevance propagated to it from neurons at layer $l+1$.

Using this method, we calculate the relevance $R_j^l$ for each neuron $j$ in layer $l$ with respect to a given input $x$. For layer $l$, assuming there are $m$ neurons with positions $p_0, \dots, p_{m-1}$ such that $p_0 < p_1 < \dots < p_{m-1}$, we create a list of pairs $(R_j^l, p_j)$ ordered by neuron position. We represent this list as:

\begin{equation}
\begin{aligned}
    List(l, x)=((R_0^l,p_0),\ldots,(R_j^l,p_j),\ldots,(R_{m-1}^l,p_{m-1})),& \\
    p_0< p_1<\ldots< p_{m-1}&
\end{aligned}
\label{List}
\end{equation}
We identify the top $k$ neurons with the highest relevance scores $R$ in $List(l, x)$ as high-relevance neurons for input $x$. By extracting these top $k$ pairs and sorting them in descending order of $R$, we obtain $List(l, x, k)$ as defined in Equation~\ref{top_k}. Unlike Equation~\ref{List}, the pairs $(R,p)$ in $List(l,x,k)$ are arranged in descending order of $R$, rather than by neuron position.

\begin{equation}
\begin{aligned}
  List(l,x,k)=((R_{0}^l,p_{0}),\ldots,(R_{k-1}^l,p_{k-1})),& \\
  \text{where } R_{0}^l\geq R_{1}^l\geq\ldots\geq R_{k-1}^l, k\in \left[ 1,m \right]&
\end{aligned}
\label{top_k}
\end{equation}
However, the analysis described above is performed for a single input sample $x$. To identify a set of neurons consistently relevant across the data associated with the unlearning class, we analyze dataset $D'$. As mentioned, $D'$ includes data samples from the unlearning class, either extracted from open-source datasets or generated synthetically to protect user privacy. Let $D' = \{x_1, \ldots, x_n\}$ be the set of $n$ data samples in $D'$. After conducting the top-$k$ relevance analysis for layer $l$ on each sample $x_j \in D'$, we calculate the occurrence frequency of each neuron position to mitigate the influence of sample-specific variations.

\begin{equation}
    C(p) = \sum_{j=1}^{n} \mathbb{I}\left( p \in \{ p \mid (R, p) \in List(l, x_j, k) \} \right)
\label{occ_frequency}
\end{equation}
As defined in Equation~\ref{occ_frequency}, $C(p)$ is the occurrence frequency of a specific neuron position $p$ across all $n$ top-$k$ relevance lists $List(l, x_j, k)$ for samples $x_j \in D'$. The indicator function $\mathbb{I}(\cdot)$ evaluates to 1 if the condition is true (i.e., if neuron position $p$ is present in the list of top-$k$ neuron positions for sample $x_j$) and 0 otherwise.

For all unique neuron positions that appeared in any of the top-$k$ lists (i.e., positions $p$ with $C(p) > 0$), we obtain their occurrence frequency $C(p)$ using the method described above. Let $P_{unique} = \{p \mid C(p) > 0\}$ be the set of these $z = |P_{unique}|$ unique neuron positions. Next, as represented in Equation~\ref{neurons_set}, we sort these neuron positions based on their frequencies $C(p)$ in descending order to identify the list of strongly relevant neurons $S_{l}$ for the unlearned class in layer $l$.

It is worth mentioning that due to sample-specific variations in relevance, the set of top-$k$ relevant neurons can vary across different samples. If we denote the number of unique neuron positions that appear in at least one of the $n$ top-$k$ lists as $z$, the value of $z$ will range between $k$ and $m$.

\begin{equation}
\begin{aligned}
    S_l &= (p_{0}, p_{1}, \ldots, p_{z-1}), \\
    &p \in P_{unique} \text{ and } C(p_{0}) \geq C(p_{1}) \geq \ldots \geq C(p_{z-1})
\end{aligned}
\label{neurons_set}
\end{equation}

\subsection{Neuronal Path Perturbation}

In neural networks for image classification, the initial layers typically extract low-level features such as edges and textures. Middle layers capture higher-level features like shapes and parts. In contrast, the FC layers transform these features into more abstract representations for classification. Therefore, we perturb the FC layer $l$ to disrupt the neuronal classification path.

First, we identify the weights connected to the neurons that need to be perturbed. These are the weights connecting the preceding layer $l-1$ to the selected neurons in layer $l$. As identified by the relevance analysis in the previous section, the set of strongly relevant neuron positions in layer $l$ for the unlearned class is $S_l$. In Equation~\ref{Weight_set}, we represent the set containing all incoming weights connected to the neurons identified in $S_l$ as $W_l$.

\begin{equation}
    W_l=\left\{ w_{0},w_{1},...,w_{z-1} \right\} ,z\in \left[ k,m \right]
\label{Weight_set}
\end{equation}

To facilitate unlearning of a specific class, we apply targeted dropout by modifying the weights connected to the neurons in layer $l$ responsible for classifying this class. As Equation~\ref{neurons_set} shows, $S_l$ is the list of strongly relevant neuron positions part of the neuronal classification path for the unlearning class in layer $l$. We implement this by setting the incoming weights to the neurons whose positions are in $S_l$ to $0$. Specifically, for each neuron position $p \in S_l$, all weights connecting from layer $l-1$ to the neuron at position $p$ in layer $l$ are set to $0$. This effectively disrupts the classification path.

This neuronal path perturbation modifies the forward propagation. We implement this by setting the weights corresponding to neurons in $S_l$ to $0$. This creates modified network parameters $\theta'$, where the weights in layer $l$ connected to neurons in $S_l$ are zeroed out. The updated prediction function is then computed using these modified parameters, denoted as $F'\left( x \right) = F\left( x, \theta' \right)$.

By applying this targeted weight masking (or dropout), the network unlearns the specific class, as the classification path through layer $l$ is effectively severed.

\subsection{Relationship Between Neuron Relevance and Unlearning}

\newtheorem{theorem}{\bf Claim}

We demonstrate that by disabling key neurons highly relevant to a specific class in a neural network, we can effectively make the network unlearn that class. To identify these neurons, we need relevance scores to analyze the weights of neurons within the neural network. This leads to the following claim:

\begin{theorem}\label{thm1}
    Setting the weights $w_p$ of neurons $p$ with high relevance scores $R_p(x_u)$ for the unlearning class $C_i$ to zero effectively achieves unlearning of class $C_i$.
\end{theorem}

This operation results in the model's prediction function $F(x; \theta)$ losing accuracy for classifying data from class $C_i$, thus realizing effective unlearning. Specifically, for an input sample $x_u$ belonging to class $C_i$, the weights are adjusted as follows:

\begin{equation}
w_p =
\begin{cases}
    0 & \text{if } R_p(x_u) \gg R_q(x_u) \text{ for all } q \notin \{ p \}_{\text{relevant}, u} \\
    w_p & \text{otherwise}
\end{cases}
\end{equation}
where $\{ p \}_{\text{relevant}, u}$ is the set of neurons with high relevance scores for the input sample $x_u$.

This adjustment modifies the model's prediction function as shown in Equation~\ref{unlearning_operation}, where $f(\cdot)$ is the activation function applied at the final layer, and $a$ represents the activation values.

\begin{equation}
\label{unlearning_operation}
\begin{aligned}
F(x; \theta') &= f\left( \sum_{\substack{p \in \{ p \}_{\text{relevant}, u} \\ R_p(x_u) \gg R_q(x_u)}} 0 \cdot a_p + \sum_{q \notin \{ p \}_{\text{relevant}, u}} w_q a_q \right) \\
&= f\left( \sum_{q \notin \{ p \}_{\text{relevant}, u}} w_q a_q \right)
\end{aligned}
\end{equation}

To prove this, consider the effect of changes in the activations of the highly relevant neurons for a given input $x$, denoted as the set $\{ p \}_{\text{relevant}, x}$, on the model output $F(x; \theta)$. The output $F(x; \theta)$ can be expressed as a function of the activations of both relevant and irrelevant neurons, as stated in Claim~\ref{thm2}.

\begin{theorem}\label{thm2}
    The output $F(x; \theta)$ is a weighted sum of the activations $a_p$ of both relevant and irrelevant neurons.
\end{theorem}

The output is then passed through an activation function $f(\cdot)$ to introduce non-linearity. Specifically, the output can be expressed as Equation~\ref{eq:output_function}, where $w_p$ and $a_p$ represent the weight and activation associated with neuron $p$.

\begin{equation}
\label{eq:output_function}
F(x; \theta) = f\left( \sum_{p \in \{ p \}_{\text{relevant}, x}} w_p a_p + \sum_{p \in \{ p \}_{\text{irrelevant}, x}} w_p a_p \right)
\end{equation}
where $\{ p \}_{\text{irrelevant}, x}$ represents the set of neurons with low relevance scores for the input $x$.

Let's now analyze the effect of perturbing the activations of the relevant neurons for a given input $x$. We first define the change in the model output as follows:

\begin{theorem}
\label{thm3}
    The change in the model output $\Delta F(x; \theta)$ is caused by the change in the activations $\Delta a_p$ of the neurons, and the magnitude of this change is determined by the sensitivity of the neuron, $\frac{\partial F(x; \theta)}{\partial a_p}$.
\end{theorem}

Based on Claim~\ref{thm2}, if we introduce a small change $\Delta a_p$ to each activation $a_p$ for $p \in \{ p \}_{\text{relevant}, x}$, the resulting change in the output $\Delta F(x; \theta)$ can be approximated using a first-order Taylor expansion around the initial activations:

\begin{align}
\label{eq:taylor_expansion}
\Delta F(x; \theta) \approx & \sum_{p \in \{ p \}_{\text{relevant, x}}} \frac{\partial F(x; \theta)}{\partial a_p} \Delta a_p \notag \\
    & + \frac{1}{2} \sum_{p, q \in \{ p \}_{\text{relevant, x}}} \frac{\partial^2 F(x; \theta)}{\partial a_p \partial a_q} \Delta a_p \Delta a_q + \cdots
\end{align}

For small perturbations $\Delta a_p$, higher-order terms can be ignored, leaving the linear approximation:

\begin{equation}
\label{eq:linear_approximation}
\Delta F(x; \theta) \approx \sum_{p \in \{ p \}_{\text{relevant}, x}} \frac{\partial F(x; \theta)}{\partial a_p} \Delta a_p
\end{equation}

where $\frac{\partial F(x; \theta)}{\partial a_p}$ is the sensitivity of the output to the activation $a_p$ of neuron $p$ for the input $x$. Since neurons in $\{ p \}_{\text{relevant}, x}$ have high relevance scores $R_p(x)$, they typically exhibit large values for $\frac{\partial F(x; \theta)}{\partial a_p}$, meaning that even small changes $\Delta a_p$ can lead to significant changes in $F(x; \theta)$.

Thus, we have:

\begin{equation}
\label{eq:output_sensitivity}
|\Delta F(x; \theta)| = \left| \sum_{p \in \{ p \}_{\text{relevant}, x}} \frac{\partial F(x; \theta)}{\partial a_p} \Delta a_p \right| \gg 0
\end{equation}

Equation~\ref{eq:output_sensitivity} demonstrates that the model’s output is highly sensitive to changes in the activations of neurons with high relevance scores $R_p(x)$ for the input $x$. Consequently, these relevant neurons play a crucial role in determining the accuracy of the model’s classification performance on the sample $x$.

\begin{theorem}
\label{thm4}
    Specific neurons store key information about a class, leading to their high relevance to the data of that class. High-relevance neurons for class $C_i$, denoted as $\{ p \}_{C_i}$, can be distinguished from low-relevance ones based on their significant relevance scores $R_p(x)$, where $R_p(x) \gg R_q(x)$ for $p \in \{ p \}_{C_i}$ and $q \notin \{ p \}_{C_i}$ for data $x$ belonging to class $C_i$.
\end{theorem}

Different neurons exhibit varying sensitivities to features, with some neurons having higher relevance scores $R_p(x)$ for key characteristics of specific categories. These neurons, $p \in \{ p \}_{C_i}$, are crucial for classification, and we may observe some overlap between categories:

\begin{equation}
\{ p \}_{C_i} \cap \{ p \}_{C_j} \neq \emptyset, \quad \forall C_i, C_j \in \mathcal{C}, \, i \neq j
\end{equation}

Therefore, identifying neurons highly relevant to the unlearning data $x_u$ is essential for machine unlearning. For a sample $x_u$ belonging to the unlearning class, let $\{ p \}_{\text{relevant}, u}$ represent the subset of neurons with high relevance scores $R_p(x_u)$. These neurons are crucial for capturing and representing the key features of $x_u$, as expressed by:

\begin{equation}
R_p(x_u) \gg R_q(x_u) \quad \text{for } p \in \{ p \}_{\text{relevant}, u}, \; q \notin \{ p \}_{\text{relevant}, u}
\end{equation}

By identifying the neurons in $\{ p \}_{\text{relevant}, u}$, we locate those primarily responsible for encoding information about the unlearning data.

Based on Claim~\ref{thm2} to \ref{thm4}, a subset of neurons with high relevance plays a key role in the classification of a specific class. Therefore, to achieve Claim~\ref{thm1}, we set the weights $w_p$ of these highly relevant neurons for the unlearning sample $x_u$ to zero, effectively removing their contributions to the model's output for that sample. Substituting $w_p = 0$ for $p \in \{ p \}_{\text{relevant}, u}$ in the output function:

\begin{align}
\label{eq:output_function_modified}
F(x; \theta') &= f\left( \sum_{p \in \{ p \}_{\text{relevant},u}} 0 \cdot a_p + \sum_{p \in \{ p \}_{\text{irrelevant},u}} w_p a_p \right) \notag \\
&= f\left( \sum_{p \in \{ p \}_{\text{irrelevant}, u}} w_p a_p \right)
\end{align}
where $\{ p \}_{\text{irrelevant}, u}$ represents the set of neurons with low relevance scores for the input sample $x_u$.

\section{Experiment}

\subsection{Datasets and Models}

We aaplied MNIST, CIFAR-10, CIFAR-100, and mini-ImageNet datasets for evaluating the performance of our method. Our experiments preprocessed the CIFAR-10, CIFAR-100, and mini-ImageNet datasets to adjust their resolutions, making them more aligned with real-world image scenarios. Table~\ref{resolutions_models} provides each dataset's image resolution and the corresponding classification model. Furthermore, to ensure that our method satisfies the definition of “Zero-shot Unlearning,” which requires the model owner to unlearn a class without accessing the training data, we extract data from the test set of each dataset as “unseen” samples for layer-wise relevance analysis.

\begin{table}[htbp]
    \centering
    \caption{Datasets with Image Resolution and Classification Model}
    \begin{tabular}{lcc}
        \toprule
        \textbf{Dataset} & \textbf{Image Resolution} & \textbf{Classification Model} \\
        \midrule
        MNIST            & 28x28    & ALLCNN     \\
        CIFAR-10         & 224x224  & ResNet-50  \\
        CIFAR-100        & 224x224  & ResNet-50  \\
        mini-ImageNet    & 256x256  & VGG-16     \\
        \bottomrule
    \end{tabular}
    \label{resolutions_models}
\end{table}

\subsection{Evaluation Metrics}

\subsubsection{Evaluation Metrics for Images unlearning}

\begin{itemize}
    \item \textbf{Unlearning class accuracy ($A_t$).} The accuracy of the model on the unlearning class. A low $A_t$ indicates successful unlearning to the target class. This metric directly quantifies the model's reduced ability to correctly classify the data points intended to be forgotten.
    \item \textbf{Global accuracy ($A_g$).} The model's average accuracy on the remaining classes is denoted as $A_g$. A high $A_g$ reflects the improved utility of the unlearned model for the remaining tasks. Maintaining a high $A_g$ ensures that the unlearning process does not significantly degrade the model's performance.
    \item \textbf{Forgetting rate ($Fr$).} The forgetting rate ($Fr$) of the unlearned model is a metric derived from a membership inference attack (MIA). This attack involves training an SVM attack model using the target model's posterior probabilities from its training data to learn how to identify unlearned class members. Subsequently, this trained SVM is used to classify the posterior probabilities produced by the unlearned model when applied to the unlearned class data. Based on this classification result, $Fr$ is calculated as the proportion of unlearned data points that this attack fails to identify as members. A higher $Fr$ value indicates more successful forgetting.
    \item \textbf{Gigabyte ($G$).} In this context, Gigabyte is employed to measure the unlearning computational resource consumption of GPU memory. We denote Gigabyte as $G$. A higher $G$ signifies a higher resource consumption. Specifically, $G$ quantifies the peak GPU memory required by the unlearning algorithm during its execution.
    \item \textbf{Time.} The time spent by the unlearning process is demonstrated as $T$. A less $T$ indicates greater efficiency. This measures the total wall-clock time required for the unlearning algorithm to complete.
\end{itemize}

\subsubsection{Comparing methods list}

\begin{itemize}
    \item \textbf{Retrain from Scratch}: Retrains the model from scratch after removing specific data, completely erasing the influence of the removed data.
    \item \textbf{Random Labels~\cite{li2023randomrelabelingefficientmachine}}: Randomizes the target class labels, making it harder for the model to recognize and learn from those data, achieving unlearning.
    \item \textbf{Amnesiac Unlearning~\cite{graves2021amnesiac}}: Tracks parameter updates during training and selectively undoes updates from batches containing sensitive data to remove its influence.
    \item \textbf{Boundary Shrink~\cite{chen2023boundary}}: Targets the decision boundary by reassigning unlearning samples to incorrect classes, shrinking the decision space of the target class.
    \item \textbf{Boundary Expanding~\cite{chen2023boundary}}: Expands the decision space by introducing a shadow class to disperse activations, then prunes it after finetuning to remove unlearning data's influence.
    \item \textbf{Neuronal Path Noise}: Introduces Gaussian (GN) or Laplacian (LN) noise to neurons identified as relevant to the unlearning dataset, disrupting the model’s ability to classify those data accurately. When applying Gaussian noise, the parameters were set to a mean of 0.0 and a standard deviation of 1.0; for Laplace noise, the parameters were set to a location of 0.0 and a scale parameter of 1.0.
\end{itemize}

\subsection{Performance Evaluation}

In image classification tasks, we apply the proposed unlearning method to remove a specific class from the MNIST, CIFAR-10, CIFAR-100, and mini-ImageNet datasets. 
Figures~\ref{MNIST:all_subfigures} to~\ref{vgg:all_subfigures} show the attribution results of LRP for unlearning and non-unlearning data before and after machine unlearning.

\subsubsection{Changes in quantitative metrics}

We use retraining as the baseline for unlearning. The retrained model’s performance on the target class serves as a reference for evaluating unlearning effectiveness. To assess the method, we apply the $\epsilon$-rule in LRP to analyze neuron relevance. In Table~\ref{performence_MNIST} to~\ref{performence_ImageNet}, for each column, the bold and underlined values represent the best and second-best results respectively. Our method delivers the most effective and well-balanced results.

For the MNIST dataset, we perform unlearning on class $1$. We extract $36$ images belonging to class $1$ and have the model classify them while analyzing the relevance patterns among its neurons.

\begin{table}[ht]
\centering
\caption{Class Unlearning Performance on the MNIST dataset.}
\label{performence_MNIST}
\begin{tabular}{cccccc}
\toprule
Method & $A_t$ & $A_g$ & $Fr$ & $G$ & $Time$\\
\midrule
Retrain from scratch        & $\textbf{0.00}$ & $\underline{0.97}$ & $\textbf{1.00}$ & $0.74$ & $71.37s$\\
Random labels               & $\textbf{0.00}$ & $\underline{0.97}$ & $\textbf{1.00}$ & $0.70$ & $15.64s$\\
Amnesiac                    & $0.38$ & $\textbf{0.98}$ & $0.78$ & $\textbf{0.44}$ & $38.72s$\\
Boundary expanding          & $0.13$ & $0.95$ & $\underline{0.91}$ & $0.65$ & $69.21s$\\
Boundary shrink             & $\underline{0.05}$ & $0.58$ & $\textbf{1.00}$ & $0.71$ & $18.55s$\\
Neuronal path noise (GN)    & $0.35$ & $0.68$ & $0.61$ & $\underline{0.64}$ & $\underline{0.77s}$\\
Neuronal path noise (LN)    & $0.21$ & $0.53$ & $0.78$ & $\underline{0.64}$ & $0.83s$\\
Neuronal path perturbation  & $\textbf{0.00}$ & $\underline{0.97}$ & $\textbf{1.00}$ & $\underline{0.64}$ & $\textbf{0.68s}$\\
\bottomrule
\end{tabular}
\end{table}

In Table~\ref{performence_MNIST}, after applying our unlearning method, the model's accuracy $A_t$ on class 1 dropped to $0.00$, matching the $A_t$ after retraining. This shows the model completely loses the ability to classify data from the unlearning class, cutting off the classification path by disrupting relevant neurons.

While removing the unlearning class path, the model's ability to classify other classes is preserved. After retraining, the model achieves an overall accuracy $A_g$ of $0.97$ on the remaining classes. With our unlearning method, $A_g$ remains $0.97$.

Our method also performs well in unlearning verification through membership inference. After retraining, the new model achieves a forgetting rate of $1.00$ for the unlearning class, indicating full unlearning. Our approach achieves the same forgetting rate, confirming that severing the classification path effectively unlearns the class.

Moreover, our method is computationally efficient. Retraining requires $71.37$ seconds and $0.74$ GPU memory, while our unlearning method completes the task in $0.68$ seconds with only $0.64$ GPU memory. These results show that our method reduces resource consumption without additional training.

\begin{table}[ht]
    \centering
    \caption{Class Unlearning Performance on the CIFAR-10 dataset}
    \label{performence_CIFAR10}
    \begin{tabular}{cccccc}
        \toprule
        Method & $A_t$ & $A_g$ & $Fr$ & $G$ & $Time$\\
        \midrule
        Retrain from scratch        & $\textbf{0.00}$ & $\underline{0.96}$ & $\textbf{1.00}$ & $6.89$ & $484.17s$ \\
        Random labels               & $\textbf{0.00}$ & $0.90$ & $\underline{0.99}$ & $6.89$ & $107.99s$ \\
        Amnesiac                    & $0.91$ & $0.84$ & $0.13$ & $\textbf{1.75}$ & $\textbf{3.30s}$ \\
        Boundary expanding          & $0.83$ & $0.89$ & $0.01$ & $4.07$ & $150.22s$ \\
        Boundary shrink             & $0.19$ & $0.58$ & $0.09$ & $6.76$ & $20.41s$ \\
        Neuronal path noise (GN)    & $0.95$ & $0.79$ & $0.00$ & $\underline{2.82}$ & $\underline{14.17s}$ \\
        Neuronal path noise (LN)    & $0.99$ & $0.71$ & $0.00$ & $\underline{2.82}$ & $15.51s$ \\
        Neuronal path perturbation  & $\underline{0.03}$ & $\textbf{0.97}$ & $\underline{0.99}$ & $\underline{2.82}$ & $14.44s$ \\
        \bottomrule
    \end{tabular}
\end{table}

The results in Table~\ref{performence_CIFAR10} show that Neuronal Path Perturbation is the most effective unlearning method for the CIFAR-10 dataset, achieving near-complete unlearning with $A_t = 0.03$, similar to retraining ($A_t = 0.00$). Its forgetting rate ($Fr = 0.99$) is close to the ideal $Fr = 1.00$, indicating strong unlearning.

At the same time, it maintains high accuracy on the remaining classes ($A_g = 0.97$), outperforming most methods and matching retraining ($A_g = 0.96$). This demonstrates its ability to preserve model utility while effectively unlearning the target class.

Additionally, Neuronal Path Perturbation is highly efficient, using only $2.82$ GB of GPU memory and taking just $14.44$ seconds, compared to retraining’s $6.89$ GB and $484.17$ seconds, making it a practical and efficient choice for unlearning.

\begin{table}[ht]
    \centering
    \caption{Class Unlearning Performance on the CIFAR-100 dataset}
    \label{performence_CIFAR100}
    \begin{tabular}{cccccc}
        \toprule
        Method & $A_t$ & $A_g$ & $Fr$ & $G$ & $Time$\\
        \midrule
        Retrain from scratch        & $\textbf{0.00}$ & $\textbf{0.82}$ & $\textbf{1.00}$ & $6.89$ & $607.89s$ \\
        Random labels               & $\textbf{0.00}$ & $\textbf{0.82}$ & $\textbf{1.00}$ & $6.89$ & $122.55s$ \\
        Amnesiac                    & $0.03$ & $0.70$ & $\underline{0.29}$ & $\textbf{1.75}$ & $18.45s$ \\
        Boundary expanding          & $0.77$ & $\underline{0.81}$ & $0.00$ & $4.07$ & $83.36s$ \\
        Boundary shrink             & $0.14$ & $0.45$ & $0.07$ & $7.02$ & $21.20s$ \\
        Neuronal path noise (GN)    & $0.21$ & $0.56$ & $0.00$ & $2.83$ & $\textbf{0.28s}$ \\
        Neuronal path noise (LN)    & $\textbf{0.00}$ & $0.29$ & $0.00$ & $2.83$ & $\underline{0.37s}$ \\
        Neuronal path perturbation  & $\underline{0.01}$ & $\textbf{0.82}$ & $\textbf{1.00}$ & $\underline{2.77}$ & $0.53s$ \\
        \bottomrule
    \end{tabular}
\end{table}

The CIFAR-100 class unlearning results show that Neuronal Path Perturbation strikes the best balance between effective unlearning and model utility. With $A_t = 0.01$ on the target class, it achieves near-complete unlearning, matching retraining and random labels ($A_t = 0.00$). Its forgetting rate ($Fr = 1.00$) fully erases target class information.

At the same time, it maintains high accuracy on remaining classes ($A_g = 0.82$), matching the retrained model. This demonstrates its ability to preserve model utility while unlearning the target class.

In terms of resource efficiency, Neuronal Path Perturbation uses only $2.77$ GB of GPU memory, far less than retraining ($6.89$ GB), and completes the process in $0.53$ seconds, compared to retraining's $607.89$ seconds. This makes it a highly efficient and practical solution for CIFAR-100 class unlearning.

\begin{table}[ht]
    \centering
    \caption{Class Unlearning Performance on the mini-ImageNet dataset}
    \label{performence_ImageNet}
    \begin{tabular}{cccccc}
        \toprule
        Method & $A_t$ & $A_g$ & $Fr$ & $G$ & $Time$\\
        \midrule
        Retrain from scratch        & $\textbf{0.00}$ & $\textbf{0.83}$ & $\textbf{1.00}$ & $6.51$ & $2865.87s$ \\
        Random labels               & $0.17$ & $0.14$ & $0.74$ & $\underline{2.47}$ & $582.32s$ \\
        Amnesiac                    & $0.53$ & $\underline{0.80}$ & $0.44$ & $7.61$ & $20.09s$ \\
        Boundary expanding          & $\underline{0.03}$ & $0.64$ & $\textbf{1.00}$ & $7.68$ & $697.89s$ \\
        Boundary shrink             & $0.35$ & $0.67$ & $\underline{0.76}$ & $9.13$ & $4.03s$ \\
        Neuronal path noise (GN)    & $0.91$ & $0.00$ & $\textbf{1.00}$ & $\textbf{0.93}$ & $\textbf{1.16s}$ \\
        Neuronal path noise (LN)    & $0.14$ & $0.00$ & $\textbf{1.00}$ & $\textbf{0.93}$ & $1.37s$ \\
        Neuronal path perturbation  & $\textbf{0.00}$ & $\textbf{0.83}$ & $\textbf{1.00}$ & $\textbf{0.93}$ & $\underline{1.35s}$ \\
        \bottomrule
    \end{tabular}
\end{table}

For the VGG-16 model pre-trained on ImageNet, we applied machine unlearning using $33$ images from the unlearning class on the mini-ImageNet dataset, which has $1,000$ classes. The goal was to make the model forget class $0$.

As shown in Table~\ref{performence_ImageNet}, after applying our method, the model’s $A_t$ dropped to $0.00$, while $A_g$ remained at $0.83$, matching the results of retraining. This confirms that our method effectively unlearned class-specific knowledge.

Verification with a membership inference attack using SVM also showed a $Fr$ of $1.00$ after both retraining and our unlearning method, indicating complete unlearning of the target class.

Retraining with a batch size of $32$ used $6.51$ GB of GPU memory, while our method only required $0.93$ GB. Additionally, our method completed unlearning in $0.68$ seconds, compared to $2865.87$ seconds for retraining, highlighting its efficiency.

\subsubsection{Changes in attribution results}

After evaluated our method with various models and datasets, we conducted attribution experiments on both the target model and the unlearned model using different data. The results show that the unlearned model exhibits more noticeable attribution changes for data in the unlearning class.

\begin{figure*}[htbp]
    \centering
    \subfigure[The attribution of data from the unlearning class \textbf{before} unlearning, using various propagation rules.]{
        \centering
        \includegraphics[width=0.45\textwidth]{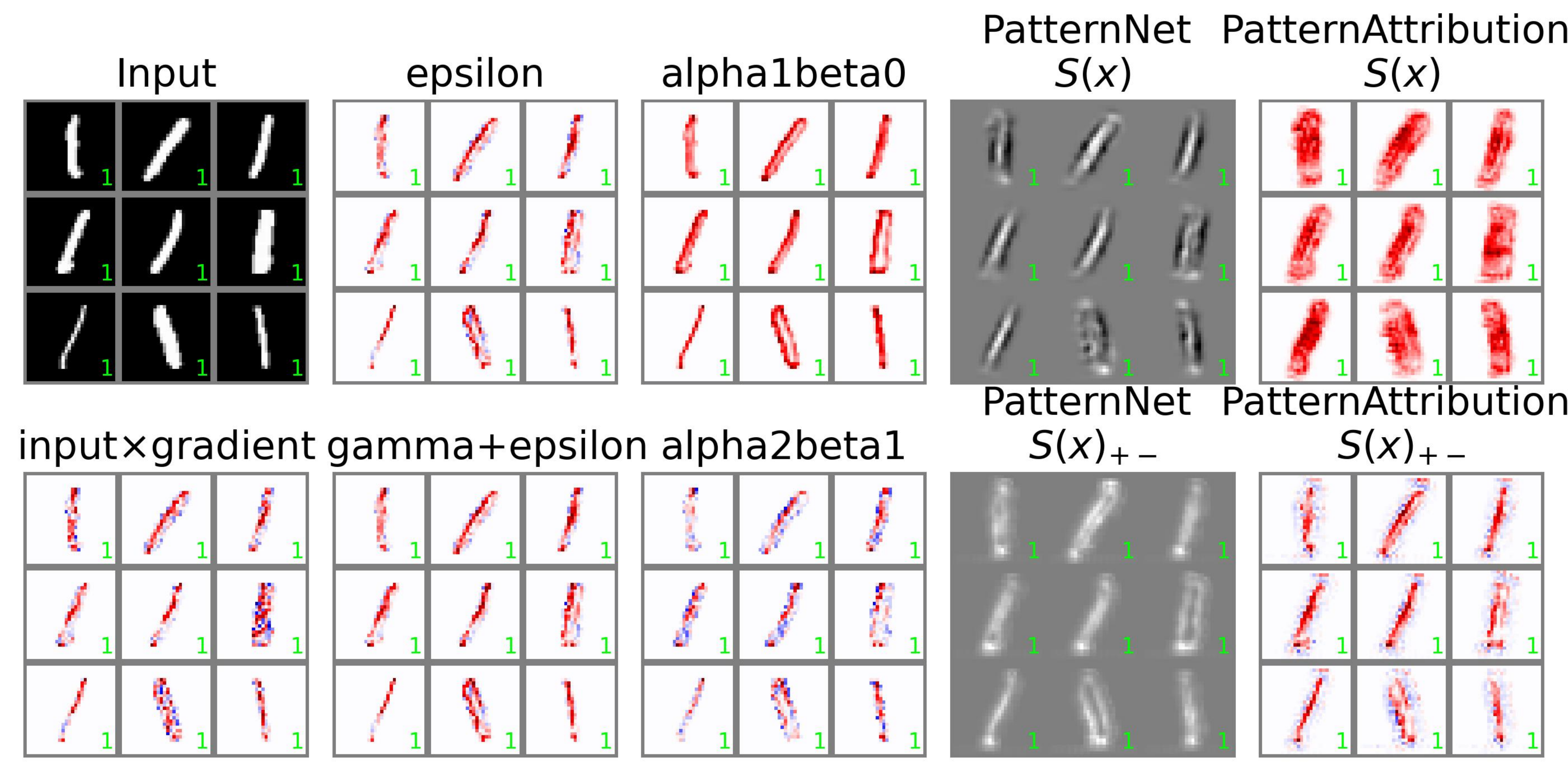}
        \label{MNIST:origin}}
    \subfigure[The attribution of data from the unlearning class \textbf{after} unlearning, using various propagation rules.]{
        \centering
        \includegraphics[width=0.45\textwidth]{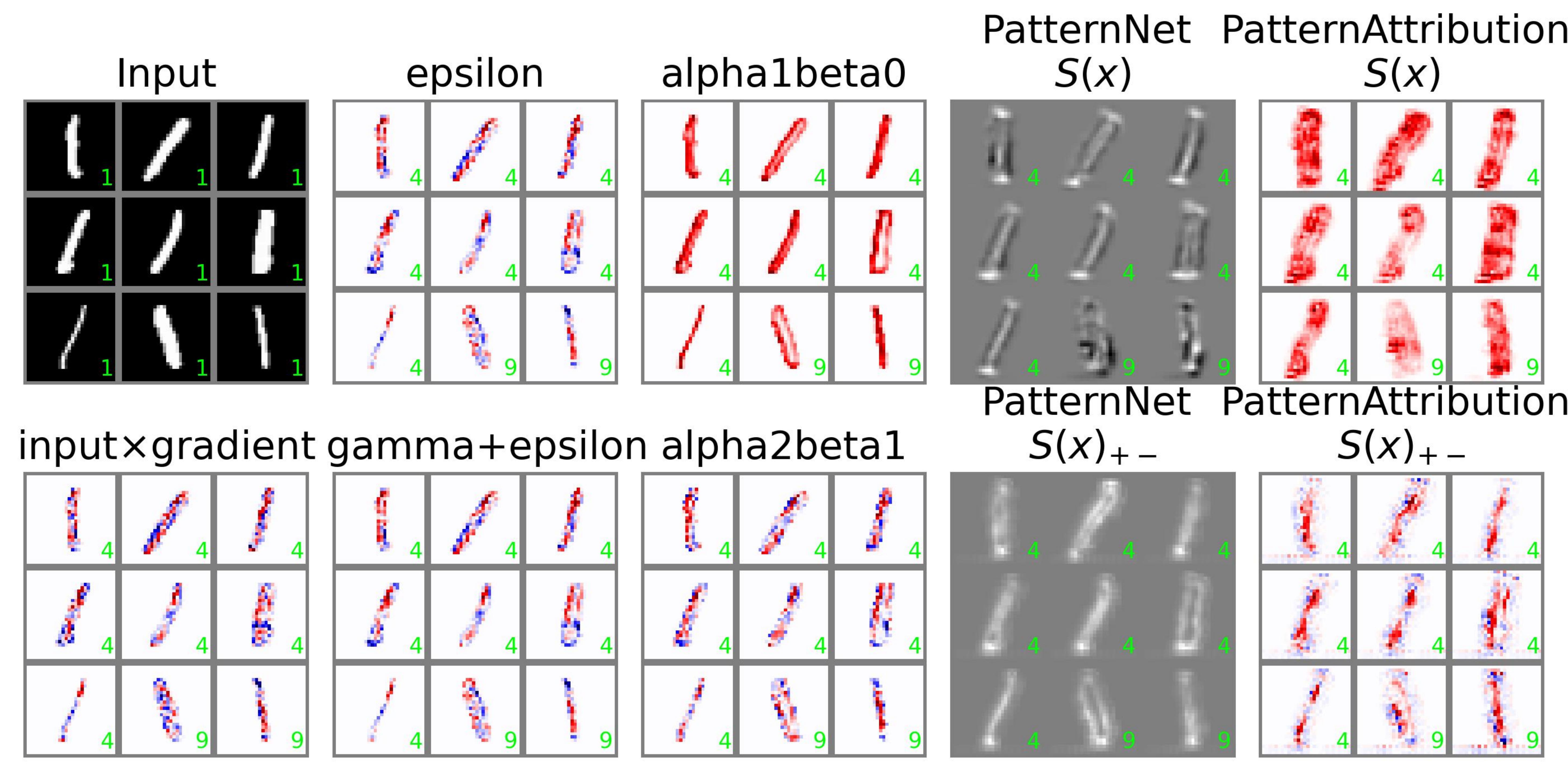}
        \label{MNIST:unlearn}}
    \subfigure[The attribution of data from the non-unlearning class \textbf{before} unlearning, using various propagation rules.]{
        \centering
        \includegraphics[width=0.45\textwidth]{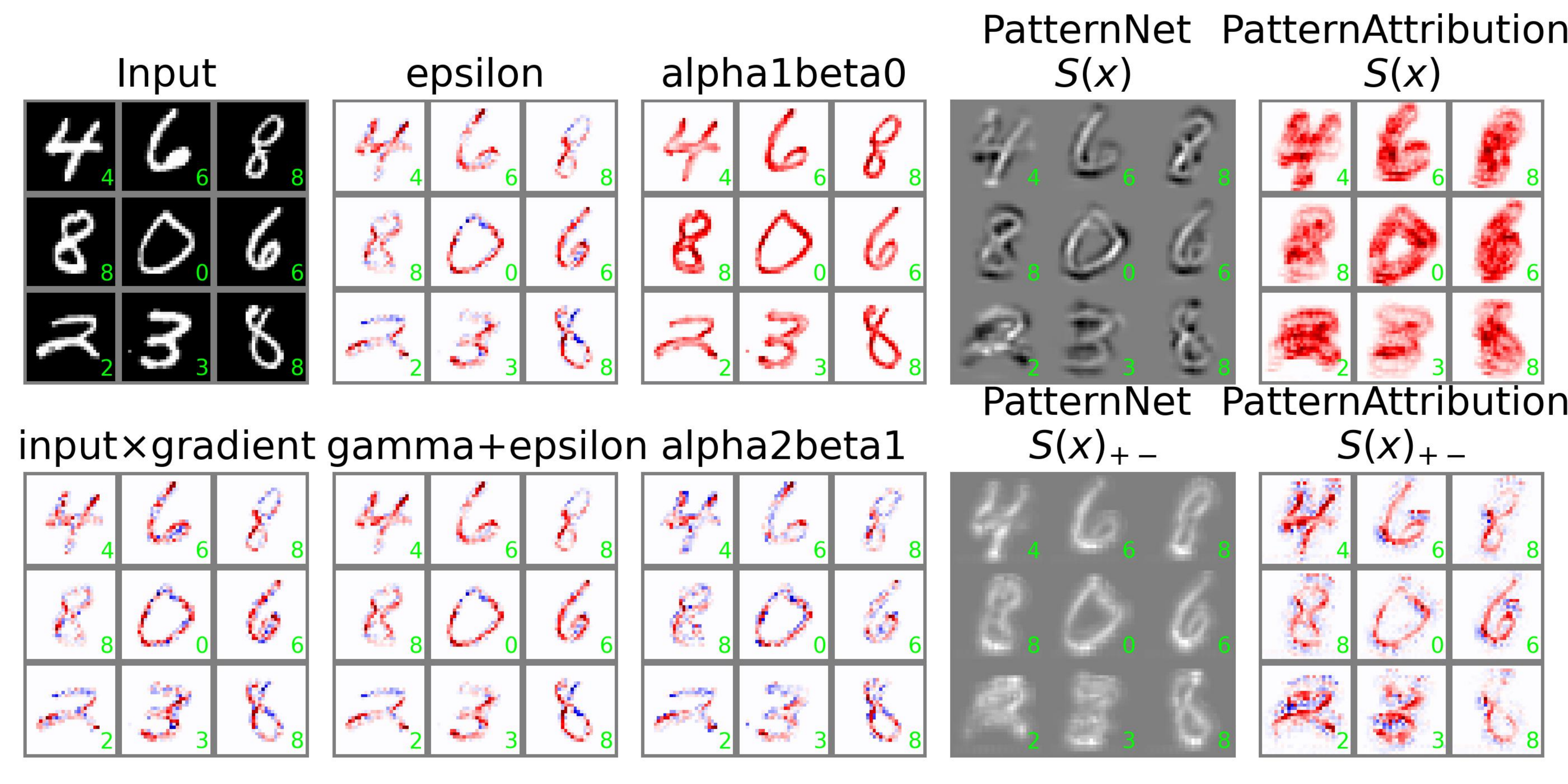}
        \label{MNIST:origin_else}}
    \subfigure[The attribution of data from the non-unlearning class \textbf{after} unlearning, using various propagation rules.]{
        \centering
        \includegraphics[width=0.45\textwidth]{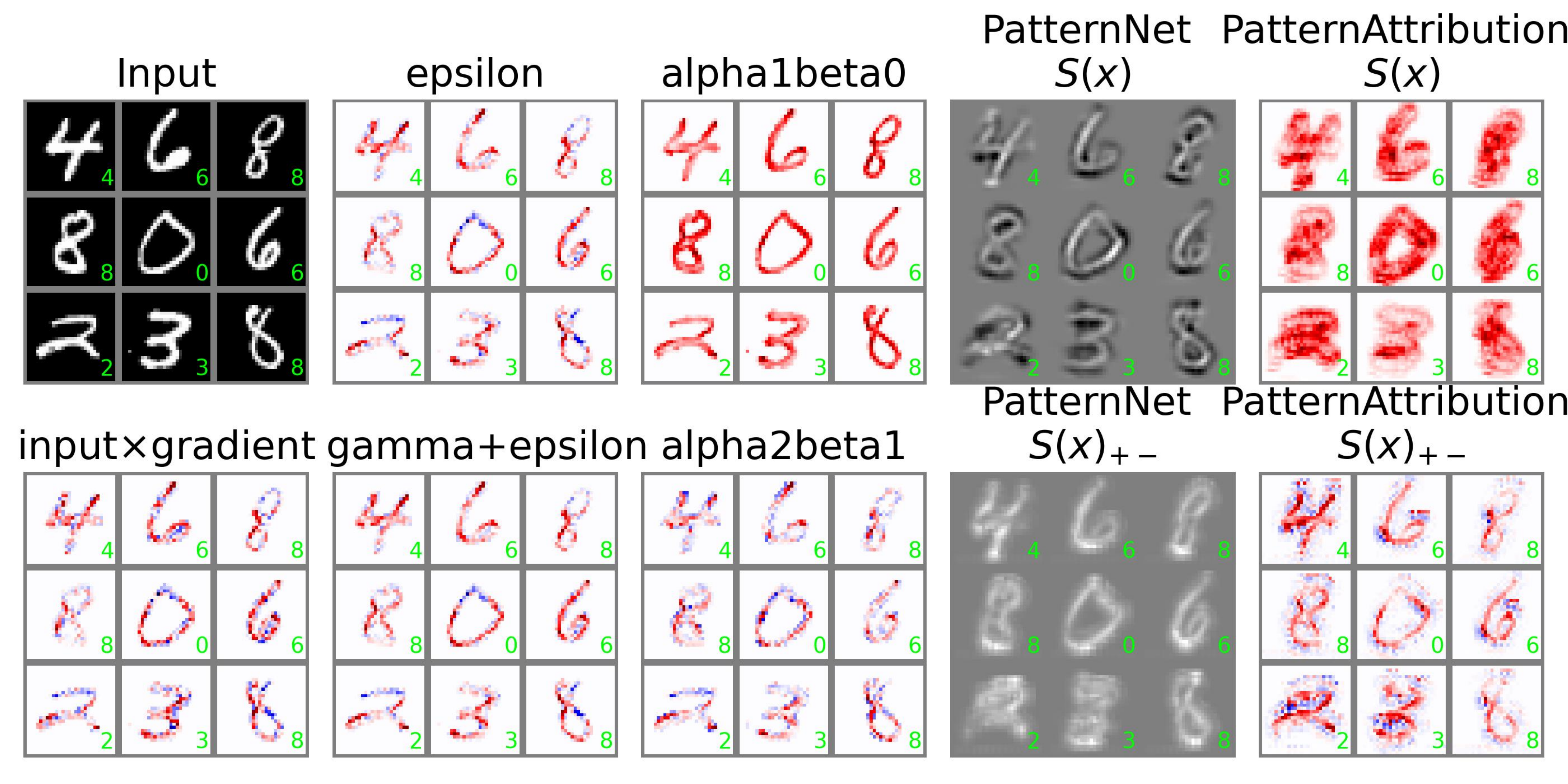}
        \label{MNIST:unlearn_else}}
    
    \caption{The attribution of the model's classification on unlearning and non-unlearning class data from \textbf{MNIST}, before and after unlearning, using different propagation rules.}
    \label{MNIST:all_subfigures}
\end{figure*}

In Figure~\ref{MNIST:all_subfigures}, the red areas represent parts of the image that positively contribute to the model's classification, while the blue areas indicate negative contributions. Specifically, Figure~\ref{MNIST:origin} and Figure~\ref{MNIST:unlearn} show the changes in attribution for the model before and after unlearning the unlearning class data.
In Figure~\ref{MNIST:unlearn}, the model still extracts the main information from the image. However, there are noticeably more pixels with negative contributions (blue pixels) compared to Figure~\ref{MNIST:origin}. This shows that the key information used for classification does not follow the original classification path in the unlearned model. Therefore, the model has successfully achieved unlearning.

In Figure~\ref{MNIST:origin_else} and Figure~\ref{MNIST:unlearn_else}, we can see that after unlearning, the model's attribution results for the same non-unlearning class data show almost no changes. Additionally, the model's classification results for the non-unlearning class remain unchanged. This indicates that our method successfully performs unlearning while preserving the model's utility on non-unlearning class data.

\begin{figure*}[htbp]
    \centering
    \subfigure[The attribution of data from the unlearning class \textbf{before} unlearning, using various propagation rules.]{
        \centering
        \includegraphics[width=0.381\textwidth]{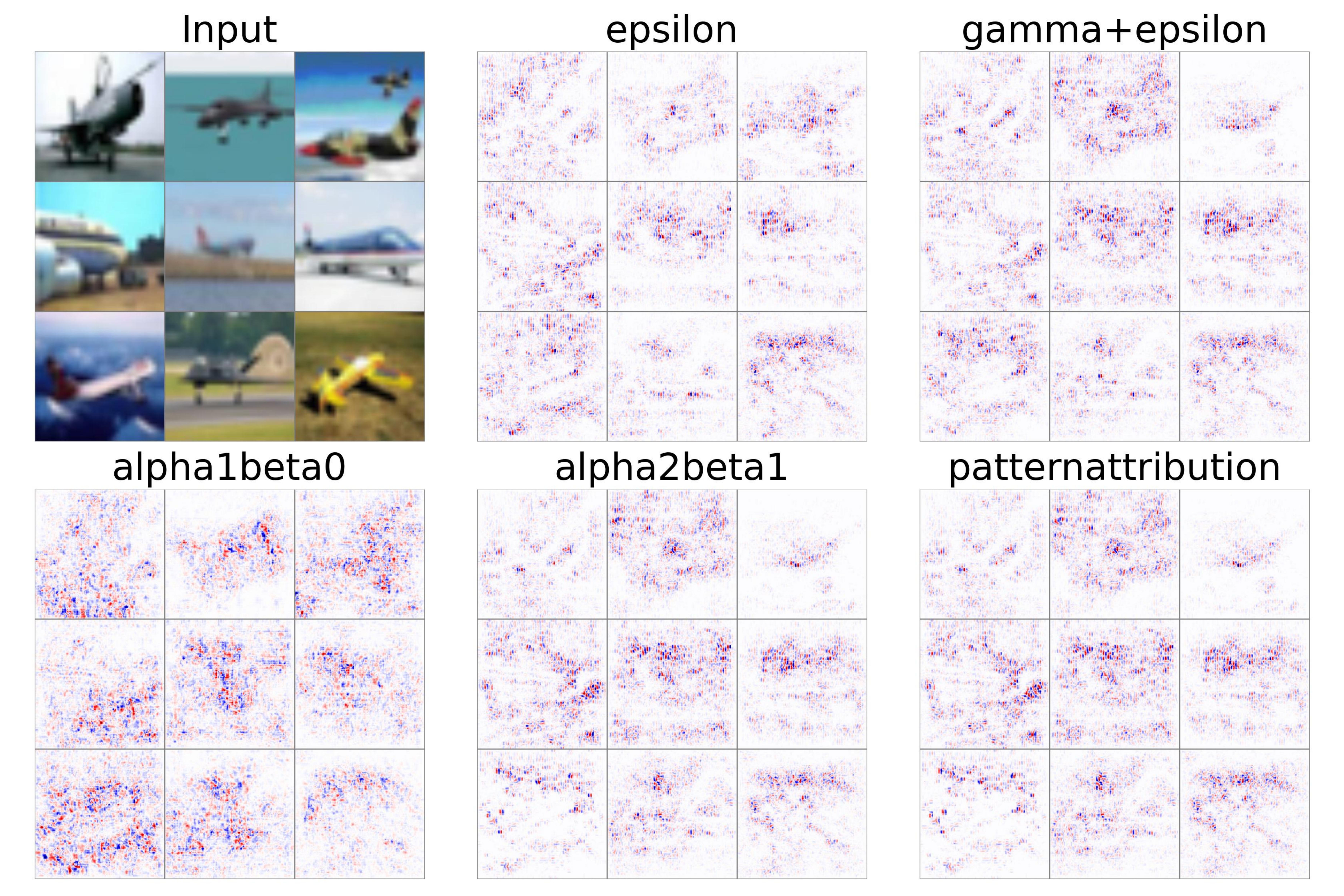}
        \label{CIFAR-10:origin}}
    \subfigure[The attribution of data from the unlearning class \textbf{after} unlearning, using various propagation rules.]{
        \centering
        \includegraphics[width=0.381\textwidth]{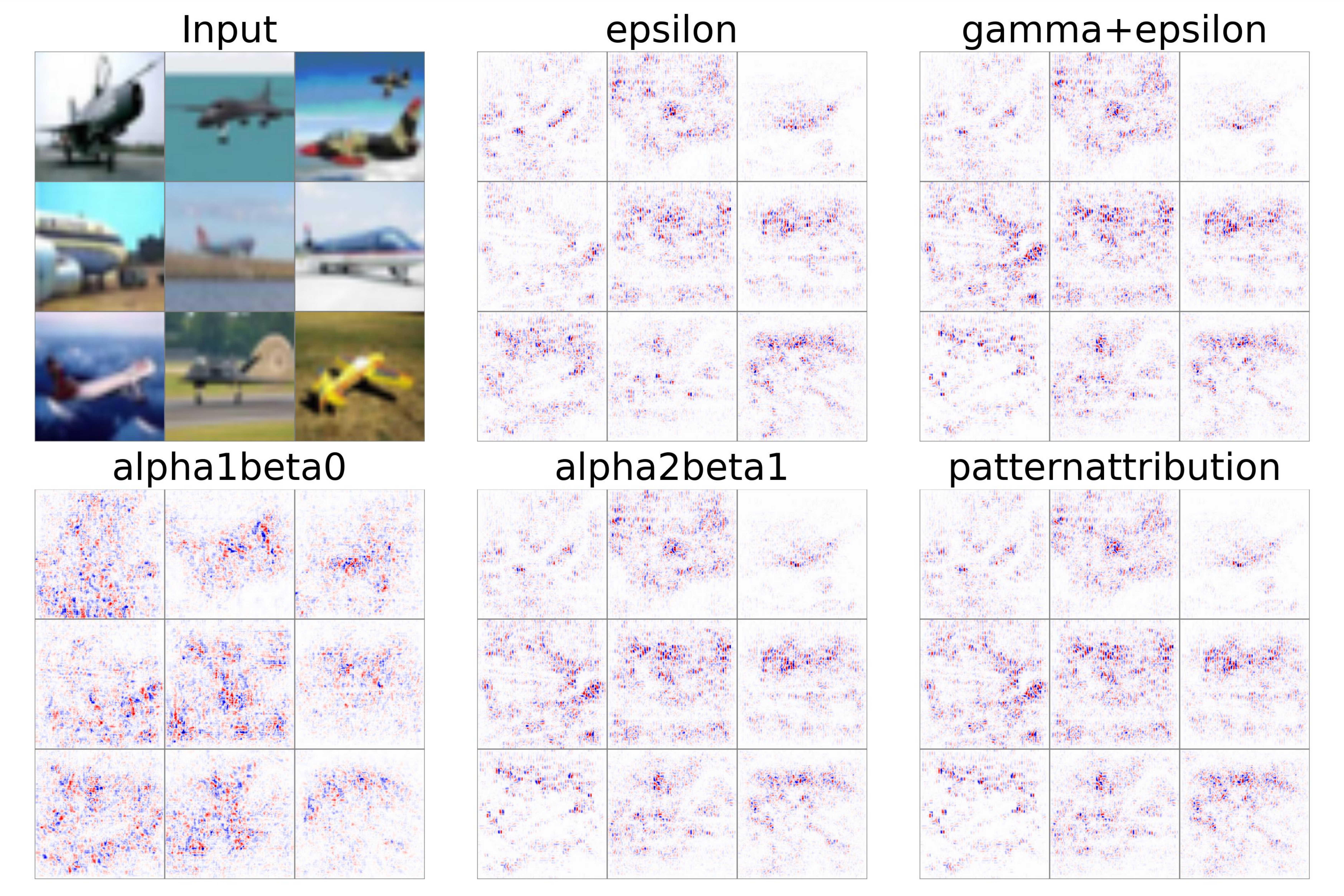}
        \label{CIFAR-10:unlearn}}
    \subfigure[The attribution of data from the non-unlearning class \textbf{after} unlearning, using various propagation rules.]{
        \centering
        \includegraphics[width=0.381\textwidth]{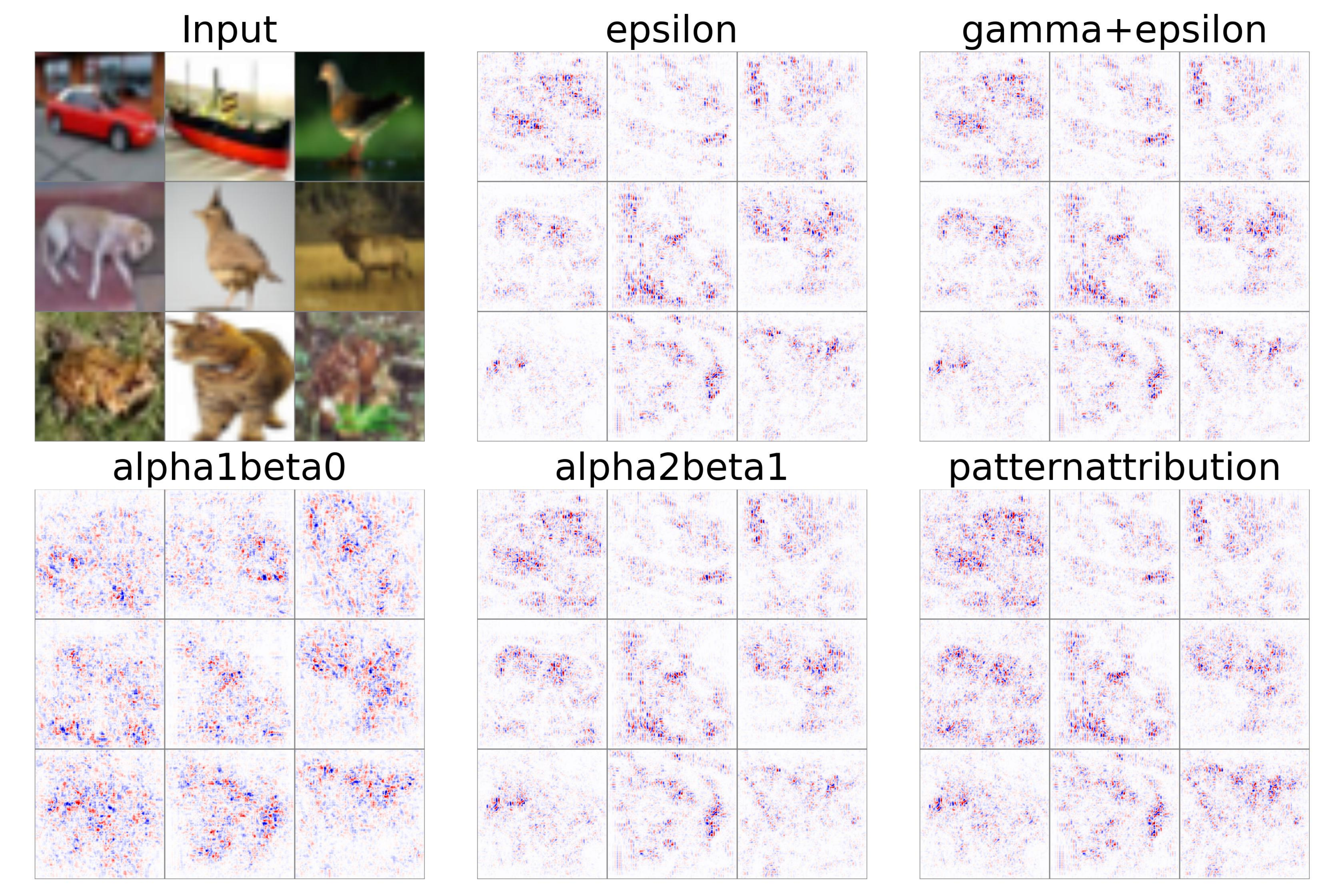}
        \label{CIFAR-10:origin_else}}
    \subfigure[The attribution of data from the non-unlearning class \textbf{after} unlearning, using various propagation rules.]{
        \centering
        \includegraphics[width=0.381\textwidth]{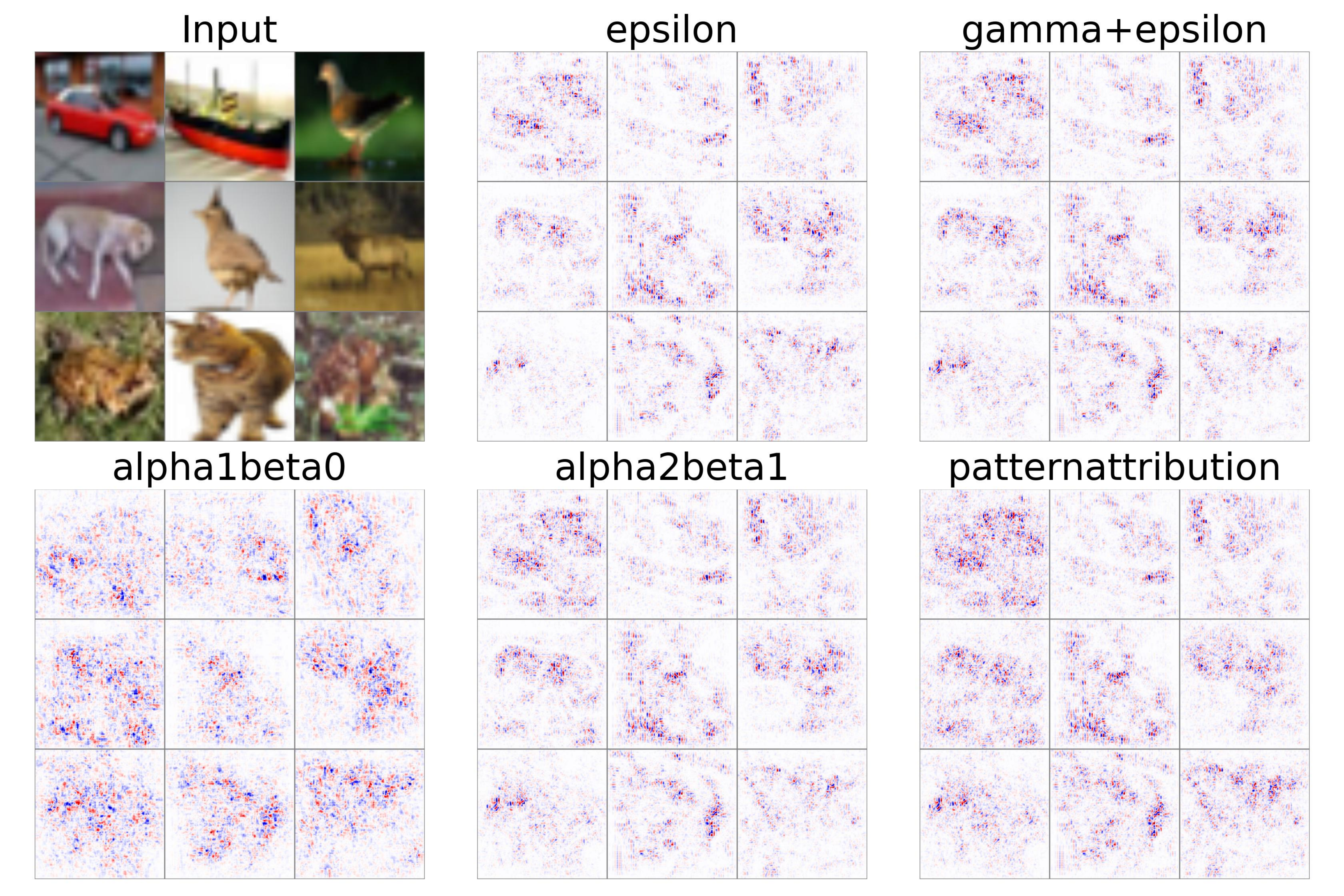}
        \label{CIFAR-10:unlearn_else}}
    
    \caption{The attribution of the model's classification on unlearning and non-unlearning class data from \textbf{CIFAR-10}, before and after unlearning, using different propagation rules.}
    \label{CIFAR-10:all_subfigures}
\end{figure*}

In Figures~\ref{CIFAR-10:origin} and~\ref{CIFAR-10:unlearn}, we compare the attributions of the ResNet50 model for unlearning class data in the CIFAR-10 dataset. Figure~\ref{CIFAR-10:origin} shows the model's attributions for unlearning class data before unlearning. After unlearning, Figure~\ref{CIFAR-10:unlearn} displays changes in the attribution distribution. In Figure~\ref{CIFAR-10:unlearn}, when using attribution rules such as $epsilon$, $gamma+epsilon$, and $alpha1beta0$, we observe that more pixels with altered attributions in the unlearning data influence the classification of the unlearned model.

Meanwhile, the unlearned model's classification on non-unlearning class data remains unaffected. In Figures~\ref{CIFAR-10:origin_else} and~\ref{CIFAR-10:unlearn_else}, we observe that the attributions for non-unlearning data show no significant differences between the original model and the unlearned model. This is because the classification paths for non-unlearning class data remain undisturbed, allowing normal forward and backward propagation. These results demonstrate that the unlearned model preserves its utility on non-unlearning data.

\begin{figure*}[htbp]
    \centering
    \subfigure[The attribution of data from the unlearning class \textbf{before} unlearning, using various propagation rules.]{
        \centering
        \includegraphics[width=0.381\textwidth]{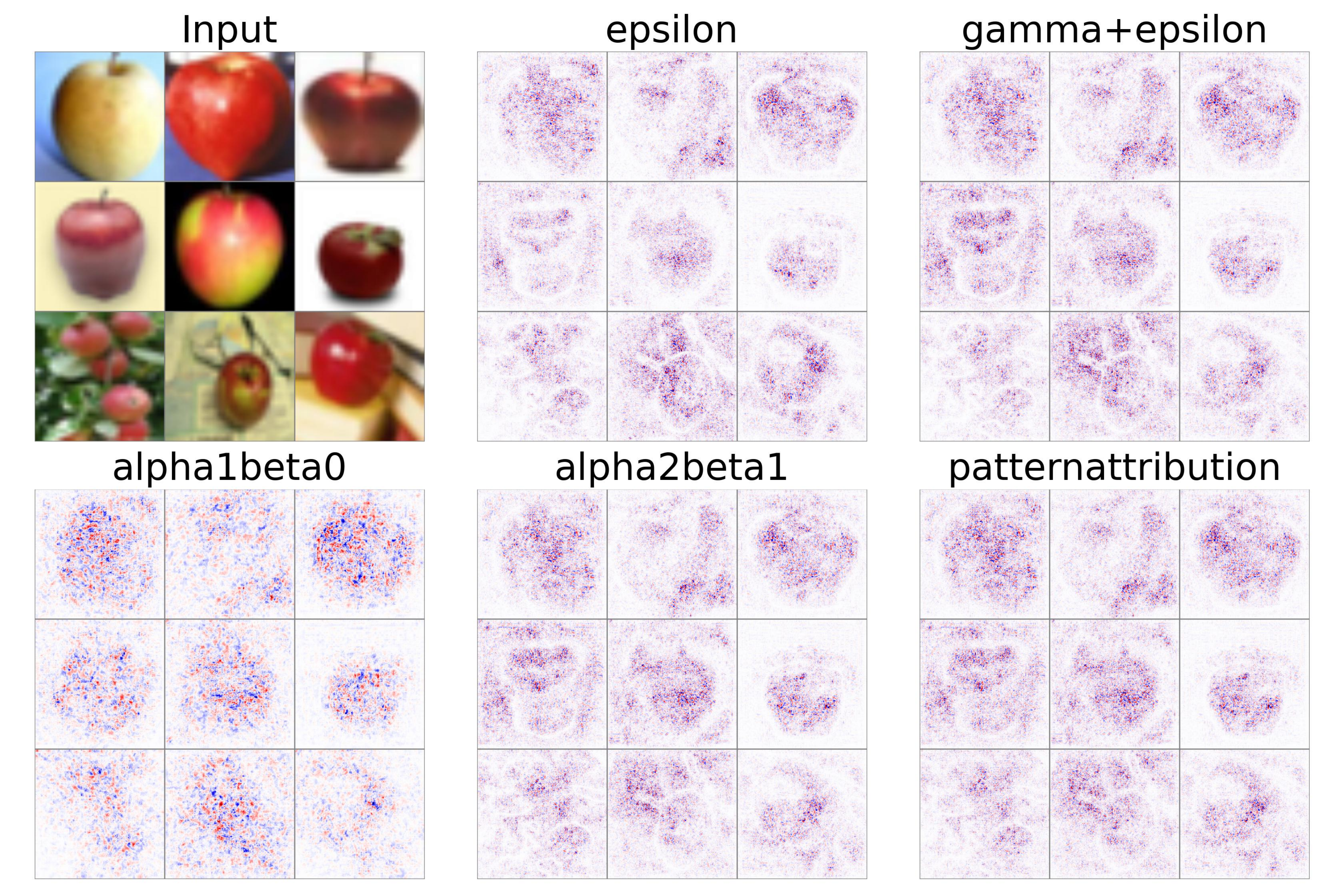}
        \label{CIFAR-100:origin}}
    \subfigure[The attribution of data from the unlearning class \textbf{after} unlearning, using various propagation rules.]{
        \centering
        \includegraphics[width=0.381\textwidth]{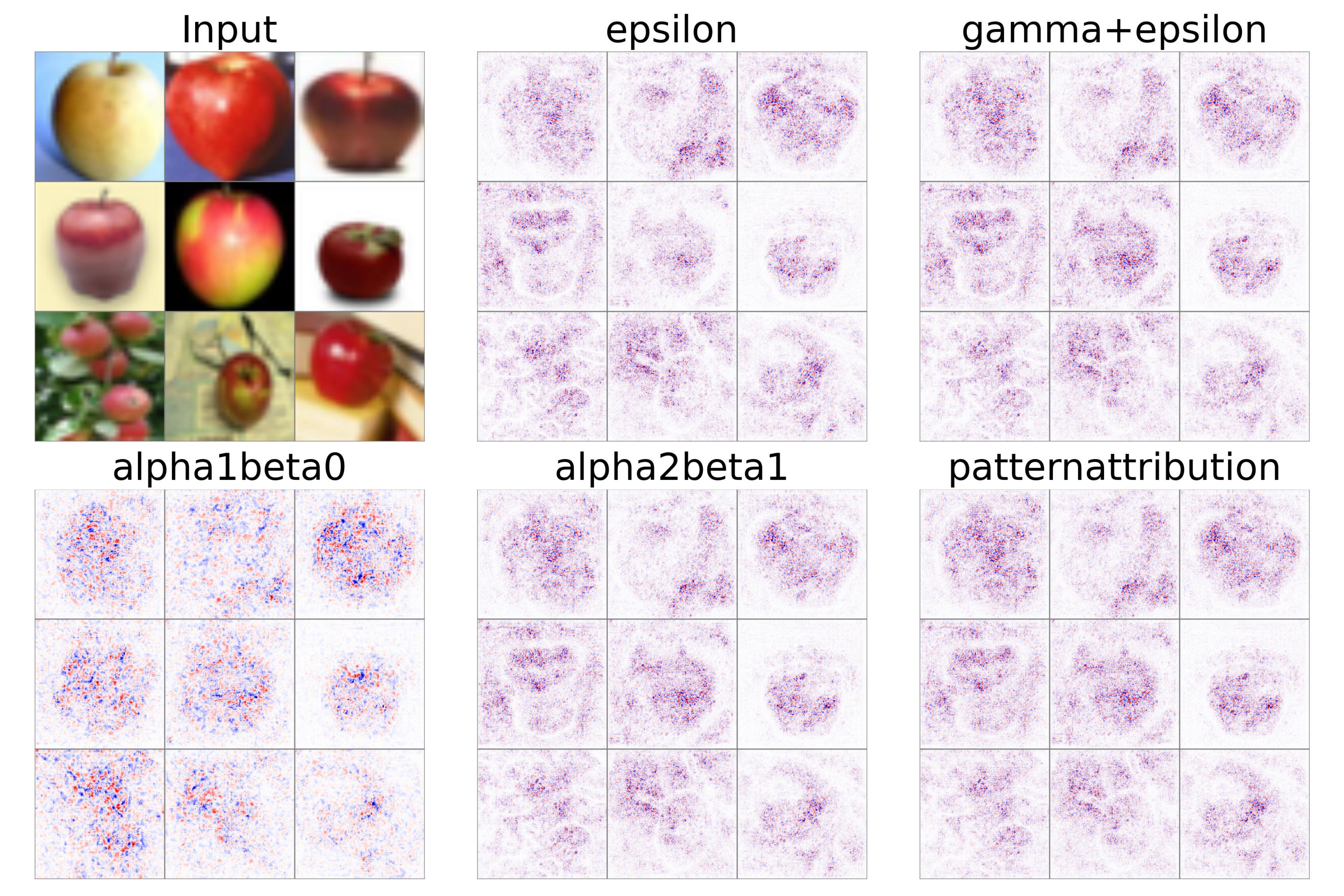}
        \label{CIFAR-100:unlearn}}
    \subfigure[The attribution of data from the non-unlearning class \textbf{after} unlearning, using various propagation rules.]{
        \centering
        \includegraphics[width=0.381\textwidth]{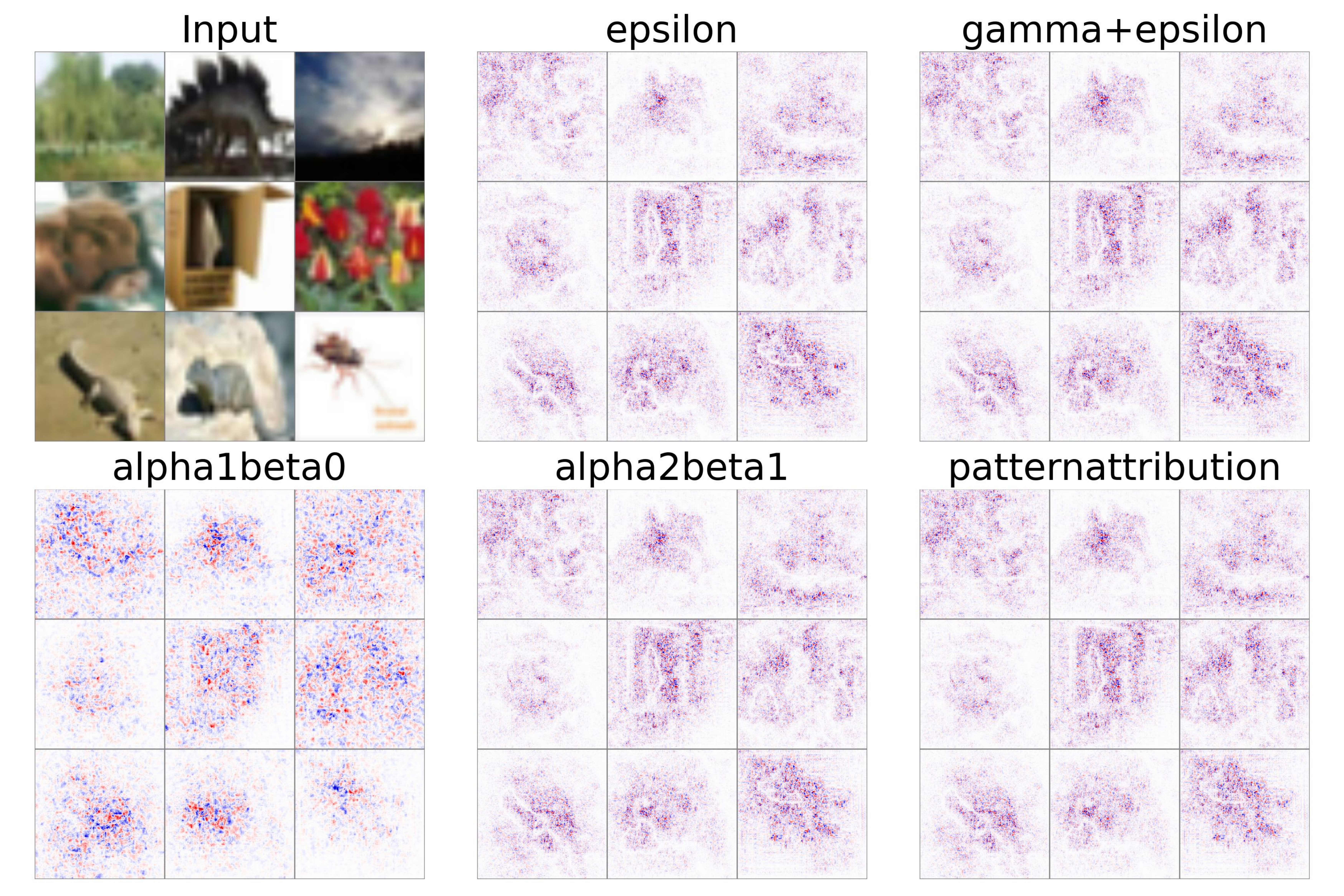}
        \label{CIFAR-100:origin_else}}
    \subfigure[The attribution of data from the non-unlearning class \textbf{after} unlearning, using various propagation rules.]{
        \centering
        \includegraphics[width=0.381\textwidth]{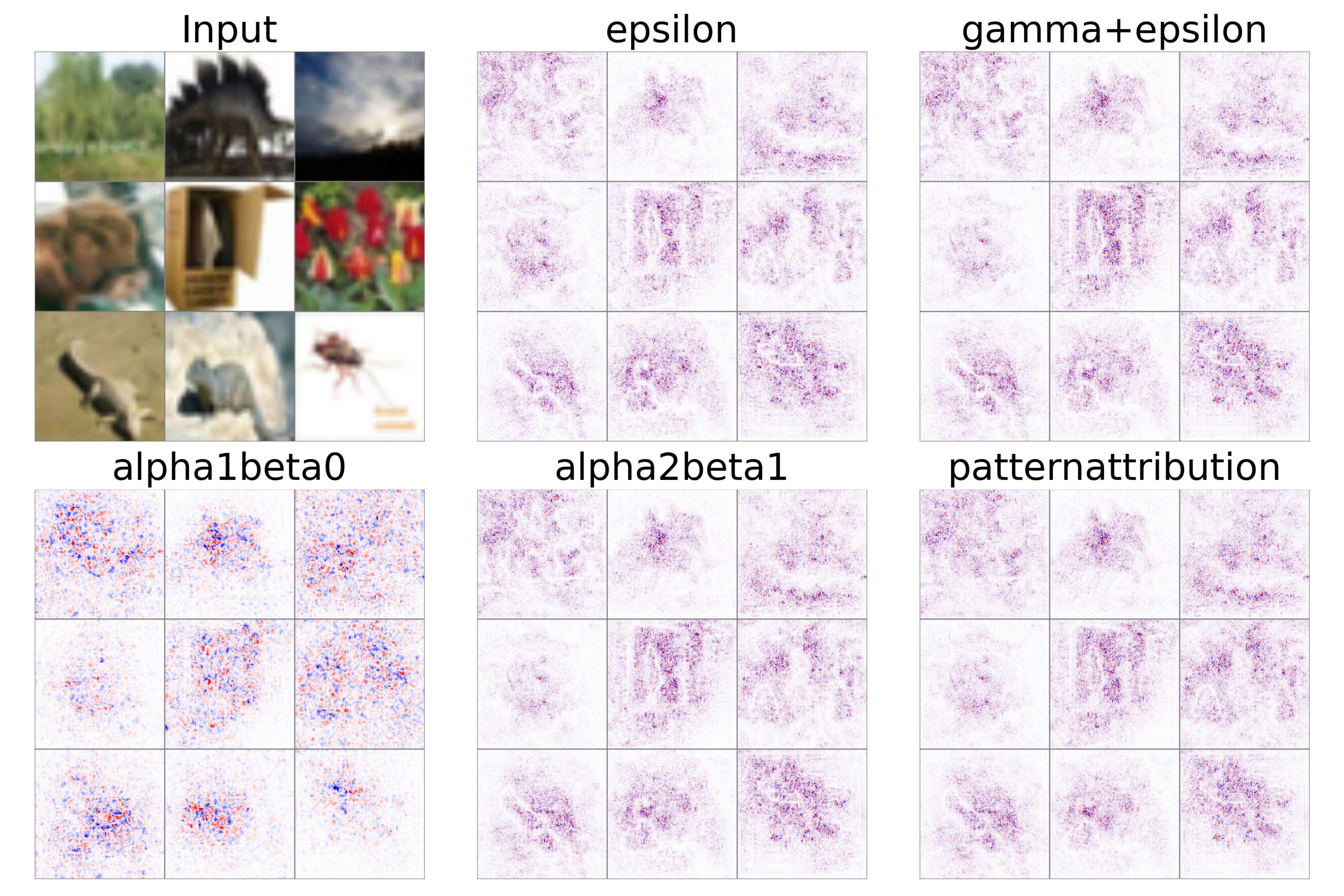}
        \label{CIFAR-100:unlearn_else}}
    
    \caption{The attribution of the model's classification on unlearning and non-unlearning class data from \textbf{CIFAR-100}, before and after unlearning, using different propagation rules.}
    \label{CIFAR-100:all_subfigures}
\end{figure*}

When performing class unlearning on data from CIFAR-100, the changes in attribution maps are relatively small, but the specified categories can still be effectively forgotten. In Figures~\ref{CIFAR-100:origin} and~\ref{CIFAR-100:unlearn}, we can see that the model extracts the main objects in the images both before and after unlearning, and the attribution results under the same rules have also changed. We believe that the minimal changes in attribution results are because, after training on a dataset with many classes, the model has lower parameter redundancy and only needs to perturb a few neurons to achieve unlearning. When a large number of neurons remain, the model can more comprehensively analyze the contribution of more pixels to image classification.

Similarly to the previous results, our unlearned model shows nearly no difference in attribution results for non-unlearning class data in CIFAR-10 compared to before unlearning. Our method effectively preserves the utility of the unlearned model.

\begin{figure*}[htbp]
    \centering
    \subfigure[The attribution of data from the unlearning class \textbf{before} unlearning, using various propagation rules.]{
        \centering
        \includegraphics[width=0.381\textwidth]{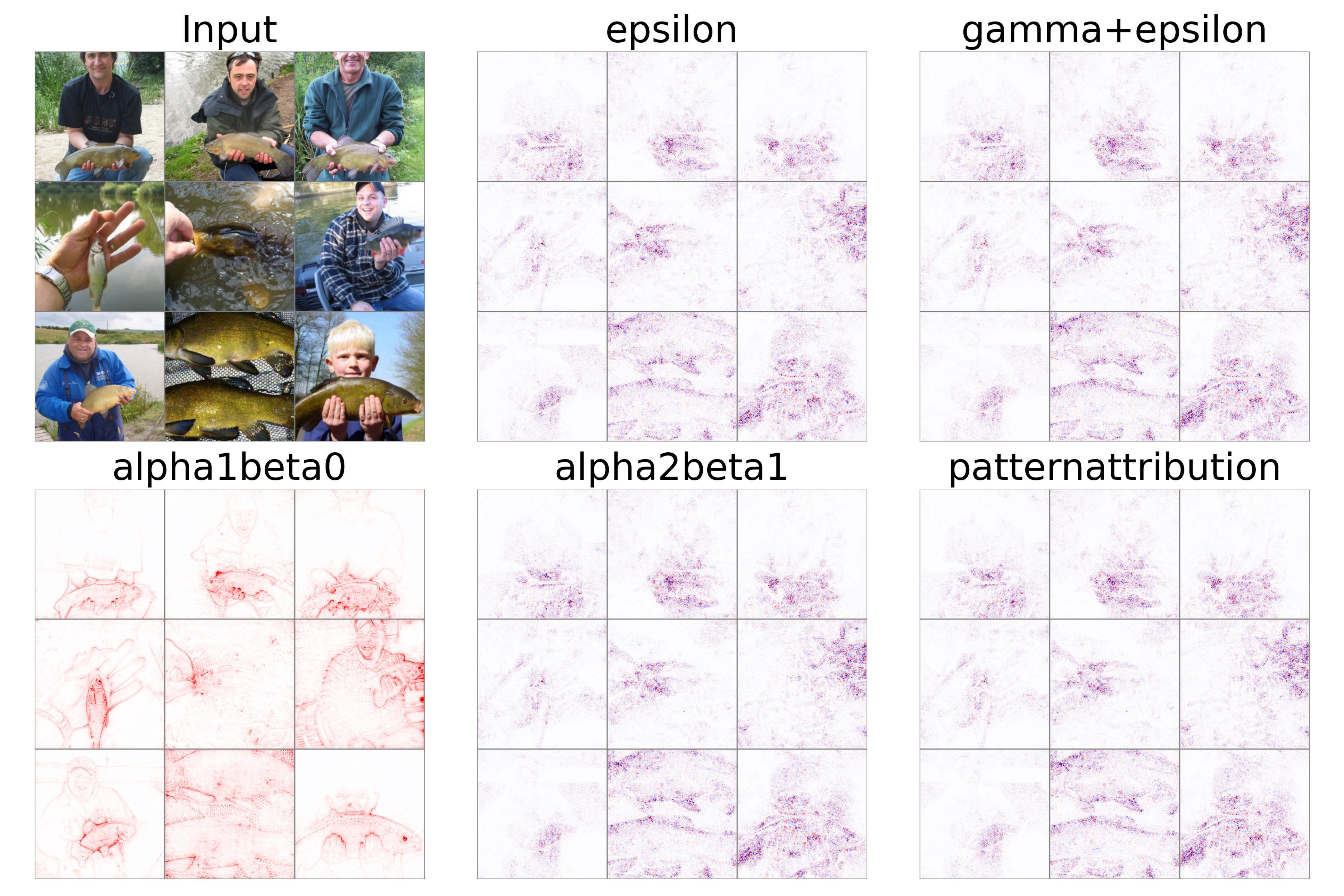}
        \label{vgg:origin}}
    \subfigure[The attribution of data from the unlearning class \textbf{after} unlearning, using various propagation rules.]{
        \centering
        \includegraphics[width=0.381\textwidth]{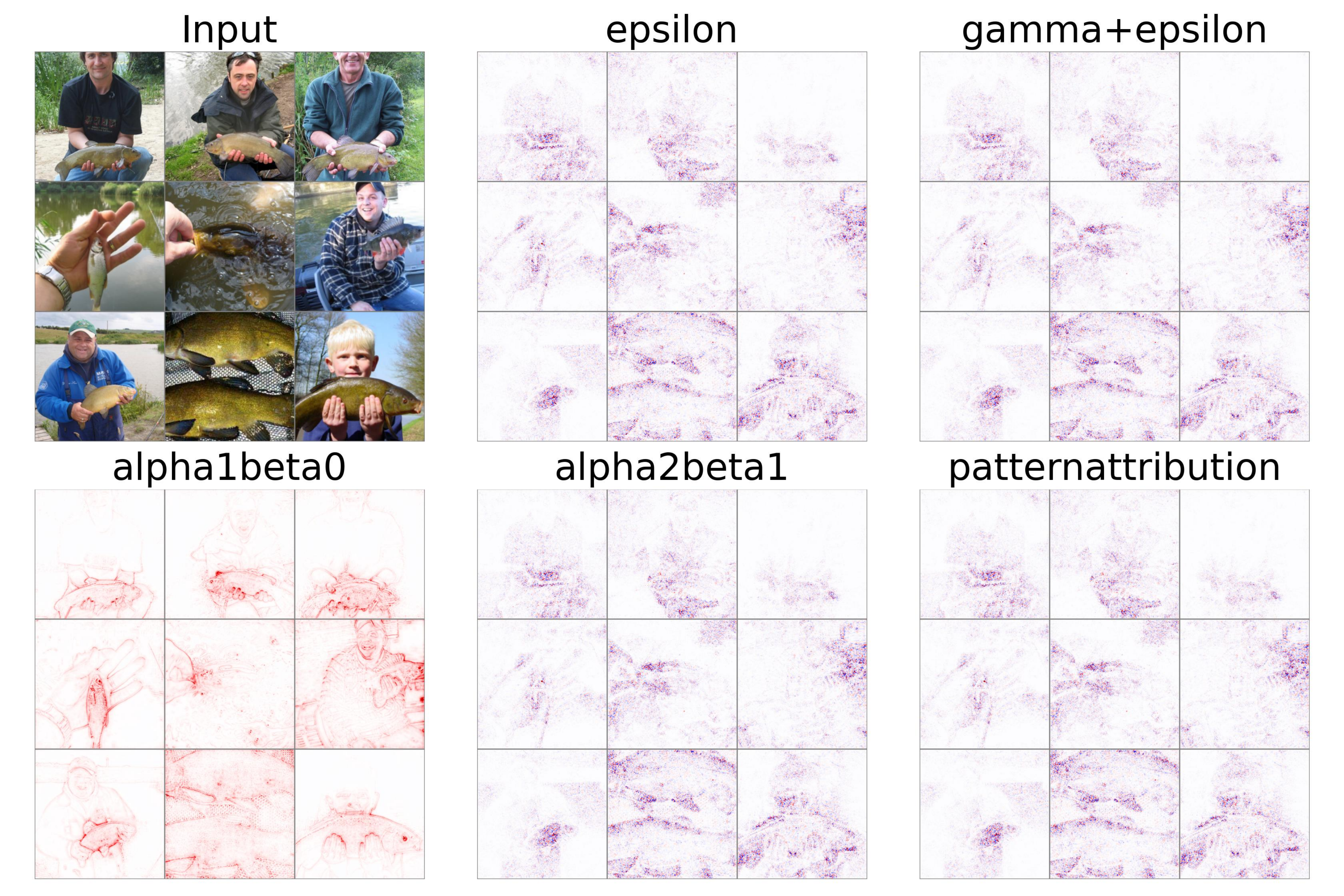}
        \label{vgg:unlearn}}
    \subfigure[The attribution of data from the non-unlearning class \textbf{after} unlearning, using various propagation rules.]{
        \centering
        \includegraphics[width=0.381\textwidth]{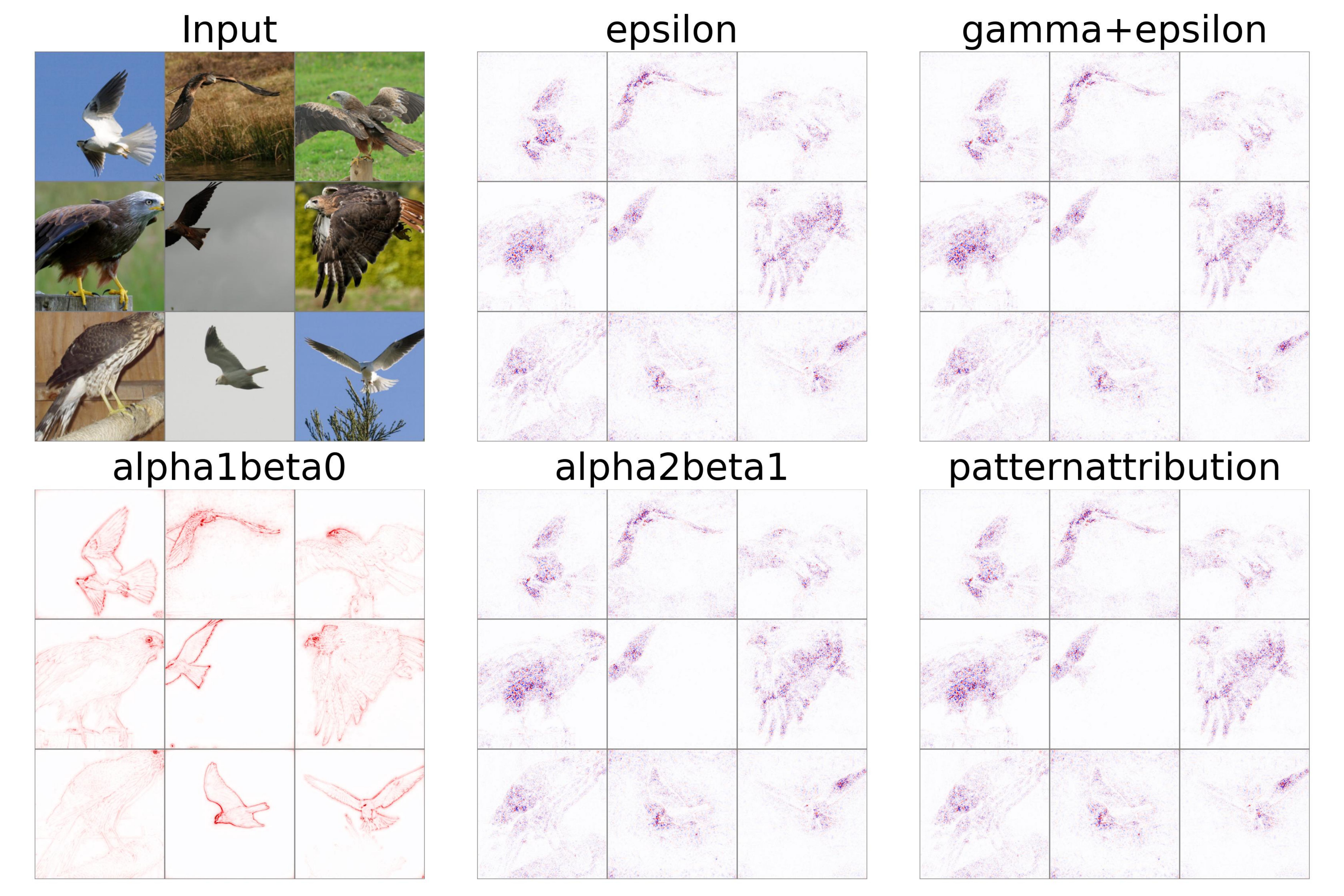}
        \label{vgg:origin_else}}
    \subfigure[The attribution of data from the non-unlearning class \textbf{after} unlearning, using various propagation rules.]{
        \centering
        \includegraphics[width=0.381\textwidth]{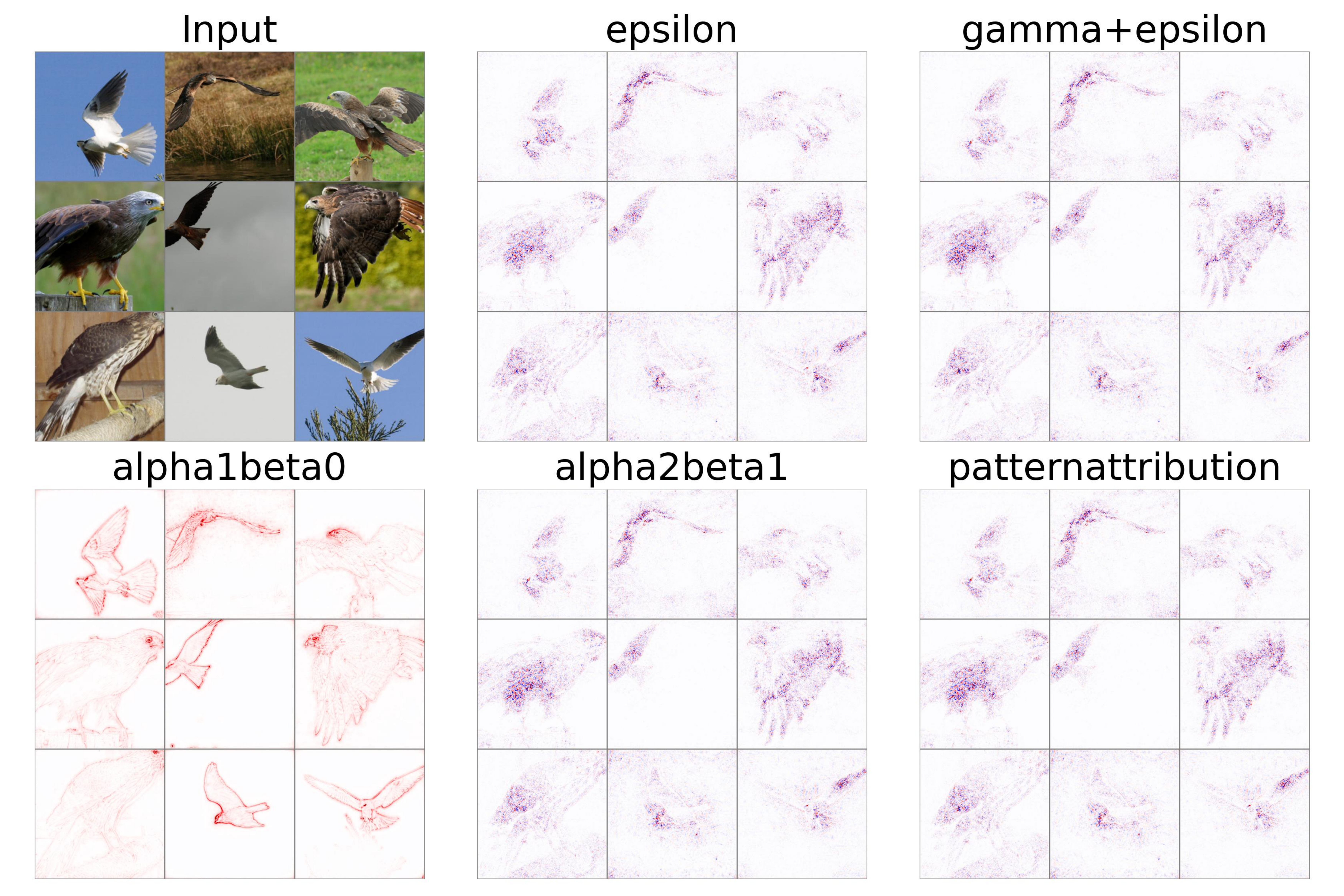}
        \label{vgg:unlearn_else}}
    
    \caption{The attribution of the model's classification on unlearning and non-unlearning class data from \textbf{mini-ImageNet}, before and after unlearning, using different propagation rules.}
    \label{vgg:all_subfigures}
\end{figure*}

In Figure~\ref{vgg:all_subfigures}, we can see the attribution results of our method validated on the mini-ImageNet dataset. Figure~\ref{vgg:origin} shows the attribution of data from the unlearning class before unlearning. After unlearning, in Figure~\ref{vgg:unlearn}, we can observe significant changes in pixel details. This is because neurons strongly related to the unlearning class data have been perturbed, causing the model to fail to classify normally.

Figure~\ref{vgg:origin_else} and Figure~\ref{vgg:unlearn_else} didn't show any difference that can be directly observed. This is because the classification paths related to non-unlearning class data didn't get perturbed during machine unlearning, indicating that our unlearning method is effective and robust.

\subsection{Parameters Evaluation}

After validating our method, we attempted to use different parameters to verify the effectiveness of our approach. We will adjust the algorithm parameters from four perspectives to validate the effectiveness and universality of our unlearning algorithm. These perspectives are: different propagation rules, different amount of perturbed neurons, different amount of analyzed neurons.

\subsubsection{Different Propagation Rules}

We use different rules to analyze and perturb different numbers of neurons to reveal the effectiveness and robustness of our approach. The relevance propagation rules that we chose are as follows:

\begin{enumerate}
    \item \emph{Epsilon}~\cite{10.1371/journal.pone.0130140} rule is one of the basic LRP rules. It helps to prevent numerical instability by ensuring that small denominators do not result in large relevance scores.
    \item \emph{Gamma+epsilon}~\cite{10.1371/journal.pone.0130140} rule is an extension of the epsilon rule, where a positive bias $\gamma$ is added to enhance the relevance of positive contributions.
    \item \emph{Alpha1beta0}~\cite{10.1371/journal.pone.0130140} rule explicitly separates the positive and negative contributions, using only the positive contributions for relevance propagation.
    \item \emph{Alpha2beta1}~\cite{10.1371/journal.pone.0130140} rule is a generalized form of the alpha1beta0 rule, where both positive and negative contributions are considered but with different weights.
    \item \emph{PatternNet $S(x)$}~\cite{kindermans2017learningexplainneuralnetworks} is designed to provide a more interpretable gradient by isolating the patterns in the input data the neural network has learned to respond to.
    \item \emph{PatternNet $S(x)_{+-}$}~\cite{kindermans2017learningexplainneuralnetworks} is a variant of PatternNet that distinguishes between positive and negative patterns, allowing for a more detailed analysis.
    \item \emph{PatternAttribution} $S(x)$~\cite{kindermans2017learningexplainneuralnetworks} extends PatternNet by attributing the network’s output prediction to the input features, similar to LRP but focusing on learned patterns.
    \item \emph{PatternAttribution} $S(x)_{+-}$~\cite{kindermans2017learningexplainneuralnetworks} rule is similar to PatternAttribution but allows for the separation of positive and negative contributions, similar to the $S(x)_{+-}$ rule in PatternNet.
\end{enumerate}

Table~\ref{mnist_table} presents the unlearning performance of our method on the MNIST dataset under various relevance propagation rules, evaluating the accuracy on the remaining data ($A_g$) and the unlearning data ($A_t$) with increasing proportions of perturbed neurons ($M_p$). Across a range of propagation rules, including \textit{epsilon}, \textit{gamma+epsilon}, \textit{alpha1beta0}, and \textit{alpha2beta1}, we consistently observe a significant reduction in $A_t$ as $M_p$ increases to $20\%$, typically dropping from an initial $0.99$ to near $0.001$. Simultaneously, these rules maintain a high $A_g$ around $0.97$, indicating a robust unlearning process that preserves performance on the unperturbed data.

While \textit{PatternNet} $S(x)$ and \textit{PatternNet} $S(x)_{+-}$ show a more gradual decrease in $A_t$ (to $0.19$ and $0.28$ respectively at $M_p = 20\%$), they still achieve substantial unlearning, albeit with a slight impact on $A_g$ (dropping to $0.95$ and $0.96$). Interestingly, \textit{PatternAttribution} $S(x)$ exhibits a less pronounced unlearning effect ($A_t$ reduces to $0.83$ at $M_p = 20\%$) while maintaining high $A_g$ ($0.97$), suggesting a trade-off between unlearning efficacy and retention. However, the modified \textit{PatternAttribution} $S(x)_{+-}$ demonstrates a sharp decline in $A_t$ to $0.00$, aligning with the performance of the epsilon-based rules.

The collective results from Table~\ref{mnist_table} underscore the versatility of our unlearning approach. Despite the diverse mechanisms of the tested relevance propagation rules, our method consistently demonstrates the ability to effectively unlearn targeted class information. The varying rates and degrees of unlearning achieved across different rules highlight the adaptability of our framework, suggesting its potential applicability across different scenarios and model architectures where different unlearning characteristics might be desired.

\begin{table*}[ht]
    \centering
    \caption{Performance of Unlearning on MNIST with\\ Varying Propagation Rules $r$, Analyzed Neuron Count $N$, and Perturbed Neuron Proportion $M_p$}
    \begin{tabular}{>{\centering\arraybackslash}m{2.5cm} >{\centering\arraybackslash}m{1cm} *{6}{>{\centering\arraybackslash}m{1.5cm}}}
        \toprule
        \renewcommand{\arraystretch}{1.2} 
        \multirow{2}{*}{$r$} & \multirow{2}{*}{Top $N$} & \multicolumn{6}{c}{$A_{g}$ / $A_{t}$ at Different $M_p~(\approx)$ Levels} \\
        \cmidrule(l){3-8}
        & & $0\%$ & $4\%$ & $8\%$ & $12\%$ & $16\%$ & $20\%$ \\
        \midrule
        \renewcommand{\arraystretch}{0.9} 
        \multirow{2}{*}{\textit{epsilon}} & 50  & $0.98$ / $0.99$ & $0.98$ / $0.99$ & $0.97$ / $0.64$ & $0.96$ / $0.13$ & $0.96$ / $0.005$ & $0.97$ / $0.001$ \\
                                          & 100 & $0.98$ / $0.99$ & $0.98$ / $0.99$ & $0.98$ / $0.99$ & $0.97$ / $0.89$ & $0.97$ / $0.34$ & $0.96$ / $0.09$ \\
        \midrule
        \multirow{2}{*}{\textit{gamma+epsilon}} & 50  & $0.98$ / $0.99$ & $0.98$ / $0.99$ & $0.98$ / $0.59$ & $0.96$ / $0.13$ & $0.96$ / $0.001$ & $0.97$ / $0.001$ \\
                                                 & 100 & $0.98$ / $0.99$ & $0.98$ / $0.99$ & $0.98$ / $0.99$ & $0.97$ / $0.89$ & $0.96$ / $0.38$ & $0.96$ / $0.09$ \\
        \midrule
        \multirow{2}{*}{\textit{alpha1beta0}} & 50  & $0.98$ / $0.99$ & $0.98$ / $0.99$ & $0.97$ / $0.56$ & $0.96$ / $0.13$ & $0.96$ / $0.003$ & $0.97$ / $0.001$ \\
                                              & 100 & $0.98$ / $0.99$ & $0.98$ / $0.99$ & $0.98$ / $0.99$ & $0.97$ / $0.89$ & $0.97$ / $0.36$ & $0.96$ / $0.09$ \\
        \midrule
        \multirow{2}{*}{\textit{alpha2beta1}} & 50  & $0.98$ / $0.99$ & $0.98$ / $0.99$ & $0.97$ / $0.61$ & $0.96$ / $0.13$ & $0.96$ / $0.007$ & $0.97$ / $0.001$ \\
                                              & 100 & $0.98$ / $0.99$ & $0.98$ / $0.99$ & $0.98$ / $0.99$ & $0.97$ / $0.89$ & $0.96$ / $0.40$ & $0.96$ / $0.09$ \\
        \midrule
        \multirow{2}{*}{\textit{PatternNet} $S(x)$} & 50  & $0.98$ / $0.99$ & $0.98$ / $0.99$ & $0.98$ / $0.95$ & $0.97$ / $0.59$ & $0.97$ / $0.38$ & $0.95$ / $0.19$ \\
                                                    & 100 & $0.98$ / $0.99$ & $0.98$ / $0.99$ & $0.98$ / $0.95$ & $0.96$ / $0.59$ & $0.96$ / $0.38$ & $0.95$ / $0.19$ \\
        \midrule
        \multirow{2}{*}{\textit{PatternNet} $S(x)_{+-}$} & 50  & $0.98$ / $0.99$ & $0.98$ / $0.99$ & $0.98$ / $0.93$ & $0.97$ / $0.62$ & $0.96$ / $0.52$ & $0.96$ / $0.28$ \\
                                                         & 100 & $0.98$ / $0.99$ & $0.98$ / $0.99$ & $0.98$ / $0.93$ & $0.97$ / $0.62$ & $0.96$ / $0.52$ & $0.96$ / $0.28$ \\
        \midrule
        \multirow{2}{*}{\textit{PatternAttribution} $S(x)$} & 50  & $0.98$ / $0.99$ & $0.98$ / $0.99$ & $0.97$ / $0.99$ & $0.97$ / $0.97$ & $0.97$ / $0.93$ & $0.97$ / $0.83$ \\
                                                            & 100 & $0.98$ / $0.99$ & $0.98$ / $0.99$ & $0.98$ / $0.99$ & $0.97$ / $0.97$ & $0.97$ / $0.93$ & $0.97$ / $0.83$ \\
        \midrule
        \multirow{2}{*}{\textit{PatternAttribution} $S(x)_{+-}$} & 50  & $0.98$ / $0.99$ & $0.98$ / $0.99$ & $0.97$ / $0.55$ & $0.96$ / $0.21$ & $0.96$ / $0.02$ & $0.95$ / $0.00$ \\
                                                                 & 100 & $0.98$ / $0.99$ & $0.98$ / $0.99$ & $0.97$ / $0.55$ & $0.96$ / $0.21$ & $0.96$ / $0.02$ & $0.95$ / $0.00$ \\
        \bottomrule
    \end{tabular}
    \label{mnist_table}
\end{table*}

Our method's efficacy extends to more complex datasets, as demonstrated by the results on CIFAR-10 (Table~\ref{Cifar10_table}). Across various layer-wise relevance propagation rules, we observe a consistent pattern of effective unlearning. Specifically, as the proportion of perturbed neurons ($M_p$) reaches $25\%$ and $30\%$, the $A_g$ remains high at approximately $0.97$, while the $A_t$ significantly drops to the range of $0.03$–$0.00$ for the epsilon, gamma+epsilon, and alpha-series rules. This indicates that our method can successfully unlearn specific information from more intricate image datasets while preserving the model's overall utility.

The robustness of our approach is further validated by experiments on the CIFAR-100 dataset (Table~\ref{cifar100_table}). Across all tested propagation rules, at $M_p = 10\%$ and $12.5\%$, the $A_g$ is consistently maintained at $0.81$, signifying stable performance on the remaining classes. Concurrently, the $A_t$ decreases substantially to $0.01$ and $0.00$, confirming the successful unlearning of the designated target class within a more fine-grained classification task.

Finally, evaluations on the challenging mini-ImageNet dataset (Table~\ref{image_net_table}) reinforce the broad applicability of our method. With $M_p$ at $10\%$ and $12\%$, all tested relevance propagation rules demonstrate effective unlearning by reducing the $A_t$ to $0.00$, while maintaining a high $A_g = 0.83$ on the non-target classes. These results across CIFAR-10, CIFAR-100, and mini-ImageNet highlight the consistent ability of our method, regardless of the specific propagation rule employed, to achieve significant unlearning of target information without compromising the model's performance on other data, underscoring its versatility and potential for widespread application.

\begin{table*}[ht]
    \centering
    \caption{Performance of Unlearning on CIFAR-10 with\\ Varying Propagation Rules $r$, Analyzed Neuron Count $N$, and Perturbed Neuron Proportion $M_p$}
    \begin{tabular}{>{\centering\arraybackslash}m{1.5cm} >{\centering\arraybackslash}m{1cm} *{6}{>{\centering\arraybackslash}m{1.5cm}}}
        \toprule
        \renewcommand{\arraystretch}{1.2} 
        \multirow{2}{*}{$r$} & \multirow{2}{*}{Top $N$} & \multicolumn{6}{c}{$A_{g}$ / $A_{t}$ at Different $M_p~(\approx)$ Levels} \\
        \cmidrule(l){3-8}
        & & $0\%$ & $10\%$ & $15\%$ & $20\%$ & $25\%$ & $30\%$ \\
        \midrule
        \renewcommand{\arraystretch}{0.9}
        \multirow{2}{*}{\textit{epsilon}} & 150 & $0.98$ / $0.99$ & $0.98$ / $0.67$ & $0.98$ / $0.34$ & $0.98$ / $0.11$ & $0.97$ / $0.03$ & $0.97$ / $0.06$ \\
        & 200 & $0.98$ / $0.99$ & $0.98$ / $0.67$ & $0.98$ / $0.33$ & $0.98$ / $0.11$ & $0.97$ / $0.01$ & $0.95$ / $0.00$ \\
        \midrule
        \multirow{2}{*}{\textit{gamma+epsilon}} & 150 & $0.98$ / $0.99$ & $0.98$ / $0.67$ & $0.98$ / $0.34$ & $0.98$ / $0.11$ & $0.97$ / $0.01$ & $0.93$ / $0.00$ \\
        & 200 & $0.98$ / $0.99$ & $0.98$ / $0.67$ & $0.98$ / $0.33$ & $0.97$ / $0.11$ & $0.97$ / $0.01$ & $0.95$ / $0.00$ \\
        \midrule
        \multirow{2}{*}{\textit{alpha1beta0}} & 150 & $0.98$ / $0.99$ & $0.98$ / $0.67$ & $0.98$ / $0.34$ & $0.98$ / $0.11$ & $0.97$ / $0.02$ & $0.93$ / $0.01$ \\
        & 200 & $0.98$ / $0.99$ & $0.98$ / $0.67$ & $0.98$ / $0.33$ & $0.98$ / $0.11$ & $0.97$ / $0.02$ & $0.96$ / $0.00$ \\
        \midrule
        \multirow{2}{*}{\textit{alpha2beta1}} & 150 & $0.98$ / $0.99$ & $0.98$ / $0.67$ & $0.98$ / $0.33$ & $0.98$ / $0.11$ & $0.97$ / $0.02$ & $0.95$ / $0.01$ \\
        & 200 & $0.98$ / $0.99$ & $0.98$ / $0.67$ & $0.98$ / $0.33$ & $0.98$ / $0.11$ & $0.97$ / $0.02$ & $0.95$ / $0.00$ \\
        \bottomrule
    \end{tabular}
    \label{Cifar10_table}
\end{table*}

\begin{table*}[ht]
    \centering
    \caption{Performance of Unlearning on CIFAR-100 with\\ Varying Propagation Rules $r$, Analyzed Neuron Count $N$, and Perturbed Neuron Proportion $M_p$}
    \setlength{\tabcolsep}{3pt} 
    \begin{tabular}{>{\centering\arraybackslash}m{2.2cm}%
                    >{\centering\arraybackslash}m{1cm}
                    *{4}{>{\centering\arraybackslash}m{1.5cm}}}
        \toprule
        \renewcommand{\arraystretch}{1.2}
        \multirow{2}{*}{$r$} & \multirow{2}{*}{Top $N$} & \multicolumn{4}{c}{$A_{g}$ / $A_{t}$ at Different $M_p~(\approx)$ Levels} \\
        \cmidrule(l){3-6}
        & & $0\%$ & $7.5\%$ & $10\%$ & $12.5\%$ \\
        \midrule
        \renewcommand{\arraystretch}{0.9}
        \multirow{2}{*}{\textit{epsilon}} & 150 & $0.82$ / $0.96$ & $0.81$ / $0.07$ & $0.81$ / $0.01$ & $0.81$ / $0.00$ \\
        & 200 & $0.82$ / $0.96$ & $0.81$ / $0.08$ & $0.81$ / $0.01$ & $0.81$ / $0.00$ \\
        \midrule
        \multirow{2}{*}{\textit{gamma+epsilon}} & 150 & $0.82$ / $0.96$ & $0.81$ / $0.07$ & $0.81$ / $0.01$ & $0.81$ / $0.00$ \\
        & 200 & $0.82$ / $0.96$ & $0.81$ / $0.10$ & $0.81$ / $0.01$ & $0.81$ / $0.00$ \\
        \midrule
        \multirow{2}{*}{\textit{alpha1beta0}} & 150 & $0.82$ / $0.96$ & $0.81$ / $0.05$ & $0.81$ / $0.01$ & $0.81$ / $0.00$ \\
        & 200 & $0.82$ / $0.96$ & $0.81$ / $0.09$ & $0.81$ / $0.01$ & $0.81$ / $0.00$ \\
        \midrule
        \multirow{2}{*}{\textit{alpha2beta1}} & 150 & $0.82$ / $0.96$ & $0.81$ / $0.07$ & $0.81$ / $0.01$ & $0.81$ / $0.00$ \\
        & 200 & $0.82$ / $0.96$ & $0.81$ / $0.09$ & $0.81$ / $0.01$ & $0.81$ / $0.00$ \\
        \bottomrule
    \end{tabular}
    \label{cifar100_table}
\end{table*}

\begin{table*}[ht]
    \centering
    \caption{Performance of Unlearning on mini-ImageNet with\\ Varying Propagation Rules $r$, Analyzed Neuron Count $N$, and Perturbed Neuron Proportion $M_p$}
    \begin{tabular}{>{\centering\arraybackslash}m{2.2cm} >{\centering\arraybackslash}m{1cm} *{6}{>{\centering\arraybackslash}m{1.5cm}}}
        \toprule
        \renewcommand{\arraystretch}{1.2} 
        \multirow{2}{*}{$r$} & \multirow{2}{*}{Top $N$} & \multicolumn{6}{c}{$A_{g}$ / $A_{t}$ at Different $M_p~(\approx)$ Levels} \\
        \cmidrule(l){3-8}
        & & $0\%$ & $2\%$ & $4\%$ & $8\%$ & $10\%$ & $12\%$ \\
        \midrule
        \renewcommand{\arraystretch}{0.9} 
        \multirow{3}{*}{\textit{epsilon}} & 100 & $0.83$ / $0.88$ & $0.83$ / $0.88$ & $0.83$ / $0.61$ & $0.83$ / $0.32$ & $0.83$ / $0.08$ & $0.83$ / $0.08$ \\
                                          & 150 & $0.83$ / $0.88$ & $0.83$ / $0.85$ & $0.83$ / $0.56$ & $0.83$ / $0.18$ & $0.83$ / $0.00$ & $0.83$ / $0.00$ \\
                                          & 200 & $0.83$ / $0.88$ & $0.83$ / $0.88$ & $0.83$ / $0.56$ & $0.83$ / $0.18$ & $0.83$ / $0.00$ & $0.83$ / $0.00$ \\
        \midrule
        \multirow{3}{*}{\textit{gamma+epsilon}} & 100 & $0.83$ / $0.88$ & $0.83$ / $0.88$ & $0.83$ / $0.62$ & $0.83$ / $0.32$ & $0.83$ / $0.08$ & $0.83$ / $0.08$ \\
                                                & 150 & $0.83$ / $0.88$ & $0.83$ / $0.85$ & $0.83$ / $0.56$ & $0.83$ / $0.18$ & $0.83$ / $0.00$ & $0.83$ / $0.00$ \\
                                                & 200 & $0.83$ / $0.88$ & $0.83$ / $0.88$ & $0.83$ / $0.56$ & $0.83$ / $0.18$ & $0.83$ / $0.00$ & $0.83$ / $0.00$ \\
        \midrule
        \multirow{3}{*}{\textit{alpha1beta0}} & 100 & $0.83$ / $0.88$ & $0.83$ / $0.88$ & $0.83$ / $0.62$ & $0.83$ / $0.32$ & $0.83$ / $0.08$ & $0.83$ / $0.08$ \\
                                              & 150 & $0.83$ / $0.88$ & $0.83$ / $0.85$ & $0.83$ / $0.56$ & $0.83$ / $0.18$ & $0.83$ / $0.00$ & $0.83$ / $0.00$ \\
                                              & 200 & $0.83$ / $0.88$ & $0.83$ / $0.88$ & $0.83$ / $0.56$ & $0.83$ / $0.18$ & $0.83$ / $0.00$ & $0.83$ / $0.00$ \\
        \midrule
        \multirow{3}{*}{\textit{alpha2beta1}} & 100 & $0.83$ / $0.88$ & $0.83$ / $0.88$ & $0.83$ / $0.62$ & $0.83$ / $0.32$ & $0.83$ / $0.08$ & $0.83$ / $0.08$ \\
                                              & 150 & $0.83$ / $0.88$ & $0.83$ / $0.85$ & $0.83$ / $0.56$ & $0.83$ / $0.18$ & $0.83$ / $0.00$ & $0.83$ / $0.00$ \\
                                              & 200 & $0.83$ / $0.88$ & $0.83$ / $0.88$ & $0.83$ / $0.56$ & $0.83$ / $0.18$ & $0.83$ / $0.00$ & $0.83$ / $0.00$ \\
        \bottomrule
    \end{tabular}
    \label{image_net_table}
\end{table*}

\subsubsection{Different Amount of Perturbed Neurons}

Our experiments, employing the epsilon rule within the layer-wise relevance propagation framework, investigated the effect of varying the proportion of perturbed neurons ($M_p$) on unlearning performance across different datasets.

Consistent across MNIST (Table~\ref{mnist_table}), CIFAR-10 (Table~\ref{Cifar10_table}), CIFAR-100 (Table~\ref{cifar100_table}), and mini-ImageNet (Table~\ref{image_net_table}), we observed a general trend: as $M_p$ increases, the target accuracy ($A_t$) steadily declines. Notably, $A_t$ eventually reaches a stable low point, beyond which further perturbation yields minimal additional unlearning. For instance, on MNIST, $A_t$ stabilizes around $0.005$ at $M_p = 16\%$, with only marginal reduction at $20\%$. Similar stabilization trends are evident on CIFAR-10 (around $M_p = 25\%-30\%$), CIFAR-100 ($M_p \geq 10\%$), and mini-ImageNet ($M_p \geq 10\%$). Throughout this process, the non-target accuracy ($A_g$) remains relatively high, indicating preserved model utility.

This behavior suggests that perturbing a small fraction of neurons does not drastically impair the model's overall memory capabilities. We attribute this resilience to the inherent redundancy in neural network representations, where information is distributed across numerous neurons and layers. Consequently, perturbing a limited number of neurons is often insufficient to overcome this redundancy, as the unperturbed neurons can compensate for the induced disruptions. Our findings support this hypothesis, as significant memory disruption, indicated by substantial and stable drops in $A_t$, typically occurs only when $M_p$ reaches a certain threshold (e.g., $\geq 16\%$ for MNIST, $\geq 10\%$ for mini-ImageNet). This demonstrates that a critical mass of perturbed neurons is required to effectively break through the distributed nature of learned information and achieve stable unlearning.

\subsubsection{Different Amount of Analyzed Neurons}

The analysis of different quantities of neurons identified as relevant to the unlearning target significantly impacts the effectiveness of the unlearning process, as demonstrated by the trends in $A_g$ and $A_t$ across varying $M_p$ and $N$.

Examining the results on MNIST (Table~\ref{mnist_table}), we observe a clear influence of the number of analyzed neurons, top $N$. For the first four rules (epsilon, gamma+epsilon, alpha1beta0, alpha2beta1), when analyzing the top 50 neurons, a substantial drop in $A_t$ occurs rapidly as $M_p$ increases beyond 8\%, reaching near zero values by $M_p=16\%$. However, when analyzing the top 100 neurons, a comparable level of $A_t$ reduction is only achieved at higher perturbation levels, typically at $M_p=20\%$. This suggests that for MNIST, increasing the number of analyzed neurons beyond a certain point ($N=50$ in this case) might include less relevant neurons. As a result, a higher $M_p$ is needed across the entire larger group of neurons to achieve the desired unlearning effect, compared to focusing on a smaller, more critical subset of neurons.

The relationship between $N$ and unlearning effectiveness appears dataset-dependent. On CIFAR-10 (Table~\ref{Cifar10_table}) and mini-ImageNet (Table~\ref{image_net_table}), increasing $N$ seems beneficial up to a certain point. For CIFAR-10, comparing $N=150$ and $N=200$ under the four rules, $N=200$ leads to a more complete reduction in $A_t$ at the highest $M_p$ values (e.g., $A_t$ reaches 0.00 at $M_p=30\%$ for $N=200$, whereas it's around 0.06 for $N=150$). Similarly, on mini-ImageNet, increasing $N$ from 100 to 150 leads to a faster and more complete $A_t$ drop, reaching 0.00 by $M_p=10\%$, compared to 0.08 for $N=100$. However, further increasing $N$ from 150 to 200 on mini-ImageNet does not yield significant additional improvement in $A_t$ reduction at high $M_p$. These findings suggest there may be an optimal number of neurons to analyze for unlearning, where increasing $N$ initially helps capture more relevant information, but exceeding this optimum introduces noise or targets neurons less crucial for the unlearning task.

On CIFAR-100 (Table~\ref{cifar100_table}), within the tested range of $N$ (150 and 200), both values exhibit very similar unlearning characteristics under all rules. The $A_t$ values drop sharply to nearly zero by $M_p=10\%$ for both $N=150$ and $N=200$, with $A_g$ remaining stable. For CIFAR-100, within this range, the number of analyzed neurons has less impact on the unlearning performance compared to the level of $M_p$.

When selecting the number of neurons to analyze, there exists a critical balance point. Analyzing too few neurons may fail to capture the truly strongly relevant neurons, resulting in incomplete unlearning. Conversely, analyzing too many neurons can introduce many irrelevant or weakly related neurons, creating noise and reducing the efficiency of the unlearning process. Therefore, finding an appropriate value of $N$—balancing the capture of key neurons and the avoidance of noise—is essential for achieving effective and efficient unlearning.

\subsection{Experiment Analysis}

\subsubsection{Impact of Neural Network Redundancy on the Number of Perturbed Neurons}

Deep learning models, particularly over-parameterized ones, often exhibit significant parameter redundancy. This means the model contains more parameters than strictly necessary to achieve a given level of performance on the training data, leading to information being distributed across many neurons and pathways. This inherent redundancy can influence how effectively specific information can be unlearned by perturbing a subset of neurons.

Our experimental results suggest that the degree of unlearning required, measured by $M_p$, is related to the apparent redundancy in the model's representation of the target class, which may vary with dataset complexity. For datasets with fewer classes, such as MNIST and CIFAR-10, where models might develop more redundant representations for each class, achieving effective unlearning typically required perturbing a higher proportion of neurons, with $A_t$ dropping significantly only at $M_p$ values ranging from 16\% to 20\% (for MNIST) or higher (for CIFAR-10).

This necessity for higher $M_p$ in seemingly more redundant representations can be attributed to the distributed nature of the learned information. When knowledge about a target class is spread across a larger number of parameters or neurons, perturbing only a small fraction of them has a limited impact on the overall representation of that class. Therefore, a larger proportion of neurons must be perturbed to effectively disrupt the redundant encoding and significantly reduce the model's ability to recognize the target class.

In contrast, for datasets with a larger number of classes, such as CIFAR-100 and mini-ImageNet, where the model might develop less redundant or more specialized representations due to the complexity of the task, effective unlearning was achieved at lower $M_p$ values. In these cases, perturbing around 10\% of neurons was often sufficient to cause a drastic drop in $A_t$. This observation supports the idea that in less redundant representations, the information might be more concentrated, making the model more sensitive to perturbations of fewer, more critical neurons.

\subsubsection{Impact of Neural Network Redundancy on the Model Utility}

\begin{table*}[ht]
    \centering
    \caption{Analysis of the Overlap of Relevant Neurons in the Perturbed Layer}
    \setlength{\tabcolsep}{3pt} 
    \begin{tabular}{cc cc cc cc cc}
        \toprule
        $D$ & Top $N$ & \multicolumn{2}{c}{$M_p~(\approx)$} & \multicolumn{2}{c}{$\mathcal{X}_{most}$@2} & \multicolumn{2}{c}{$\mathcal{X}$@10} & \multicolumn{2}{c}{$Share$@$most$} \\
        \cmidrule(lr){3-4} \cmidrule(lr){5-6} \cmidrule(lr){7-8} \cmidrule(lr){9-10}
        & & Proportion & Quantity & Proportion & Quantity & Proportion & Quantity & Quantity & Position \\
        \midrule
        MNIST         & 50  & 0.16 & 80  & 0.43 & 34  & 0 & 0 & 6 & [155, 345] \\
        CIFAR-10      & 200 & 0.20 & 400 & 0.45 & 178 & 0 & 0 & 7 & [1447] \\
        CIFAR-100     & 150 & 0.10 & 200 & 0.30 & 60  & 0 & 0 & 5 & [1064] \\
        mini-ImageNet & 150 & 0.10 & 400 & 0.47 & 189 & 0 & 0 & 9 & [3452] \\
        \bottomrule
    \end{tabular}
    \label{tab:overlap_neurons}
\end{table*}

A key finding from our experiments is that the neuron perturbation unlearning methods generally have minimal impact on the model's overall utility, as measured by $A_g$. This robustness of $A_g$ during targeted unlearning is largely a consequence of the inherent redundancy present in deep neural networks. In these over-parameterized models, data inference and feature extraction are distributed across numerous layers, neurons, and intricate pathways~\cite{DBLP:conf/cvpr/BauZKO017,DBLP:conf/cvpr/GirishMGCDS21}.

We conducted experiments on the layers of different neural networks trained on four datasets to analyze the distribution of class-related information. We selected the top ten classes from four datasets and conducted experimental analysis under the condition of minimizing the parameters required for unlearning. We defined the following metrics to analyze the information overlap between neurons:

\begin{itemize}
    \item $\mathcal{X}_{most}$@2 represents the maximum number of shared neurons among strongly relevant neurons in the perturbed FC layer between any two classes. The more shared relevant neurons required to store information for both classes, the more likely unlearning one class will affect the model's classification utility for the other class.
    \item $\mathcal{X}$@10 represents the number of shared neurons in the FC layer perturbed by our method that exhibit strong relevance with the classification of ten classes. A higher value indicates that path perturbation on one class is more likely to adversely affect the model's classification utility for other classes.
    \item $Share$@$most$ indicates the maximum number of classes whose classifications are correlated with a single neuron among the strongly correlated neurons we analyzed.
\end{itemize}

In Table~\ref{tab:overlap_neurons}, we observe that the overlap of neurons strongly relevant to two different classes in a single FC layer of neural networks trained on four datasets, $\mathcal{X}_{most}$@2, is less than $0.5$. This phenomenon indicates that, although neurons in a single layer do share information for certain classes, each class has its own strongly correlated neurons, allowing perturbation of one class's classification path without affecting the model's utility for other classes.

Furthermore, we aimed to verify whether any neurons are strongly relevant to all ten classes. As shown in Table~\ref{tab:overlap_neurons}, the experimental results indicate that $\mathcal{X}$@10 is $0$ for all models upon completing unlearning, meaning no single neuron simultaneously serves the classification of these ten classes. This further confirms the specific relevance of neurons with different class data.

Does a single fully connected layer (FC layer) in a neural network contain neurons strongly relevant to multiple classes? We further analyzed this by calculating the $Share$@$most$ for each model. We observed that, when analyzing the strongly relevant neurons for the ten classes in the MNIST dataset, only two neurons (No. $135$ and No. $345$) were shared by a maximum of six classes. In two ResNet-50 models trained on CIFAR-10 and CIFAR-100, only one neuron each (No. $1447$ and No. $1064$) was shared by a maximum of seven and five classes, respectively. This result suggests that the knowledge learned for each class is represented by specific strongly relevant neurons forming distinct classification paths. For VGG-16 trained on the mini-ImageNet dataset, which has a single layer with $4096$ neurons, we calculated the top 400 strongly relevant neurons for each of the ten classes, but found only one neuron (No. 3452) serving a maximum of nine classes’ classification tasks.

The low neuron overlap, absence of universally critical neurons demonstrate that neural networks have sufficient redundancy to maintain distinct classification paths for each class. Due to this redundancy, no single neuron or a small set of neurons is solely responsible for the model's entire performance or the inference for general classes. Instead, the model's ability to generalize relies on the collective computation and interaction of a vast number of parameters. Consequently, perturbing or modifying a limited subset of neurons specifically identified for unlearning the target class does not cause a catastrophic breakdown of the network's fundamental functionality or its ability to correctly classify examples from other classes.

\section{Conclusion}

In this paper, we present a novel approach to machine unlearning via Layer-wise Relevance Analysis and Neuronal Path Perturbation. Layer-wise Relevance Analysis provides a detailed understanding of the internal workings of the neural network during the unlearning process. Assessing the relevance of each neuron to the unlearning information allows for more precise identification of the neurons that need to be perturbed. Neuronal Path Perturbation further enhances the unlearning process. By perturbing only the highly relevant neurons, it effectively disrupts the model's memory of the specific data without overwriting or unnecessarily modifying a large portion of the network. Compared with traditional methods, our method addresses critical challenges of machine unlearning methods, such as lack of explanation, privacy guarantees, and the balance between unlearning effectiveness and model utility. 

\ifCLASSOPTIONcaptionsoff
  \newpage
\fi

\bibliographystyle{IEEEtran}
\bibliography{IEEEabrv,ref.bib}

\begin{IEEEbiography}
[{\includegraphics[width=1in,height=1.25in,clip,keepaspectratio]{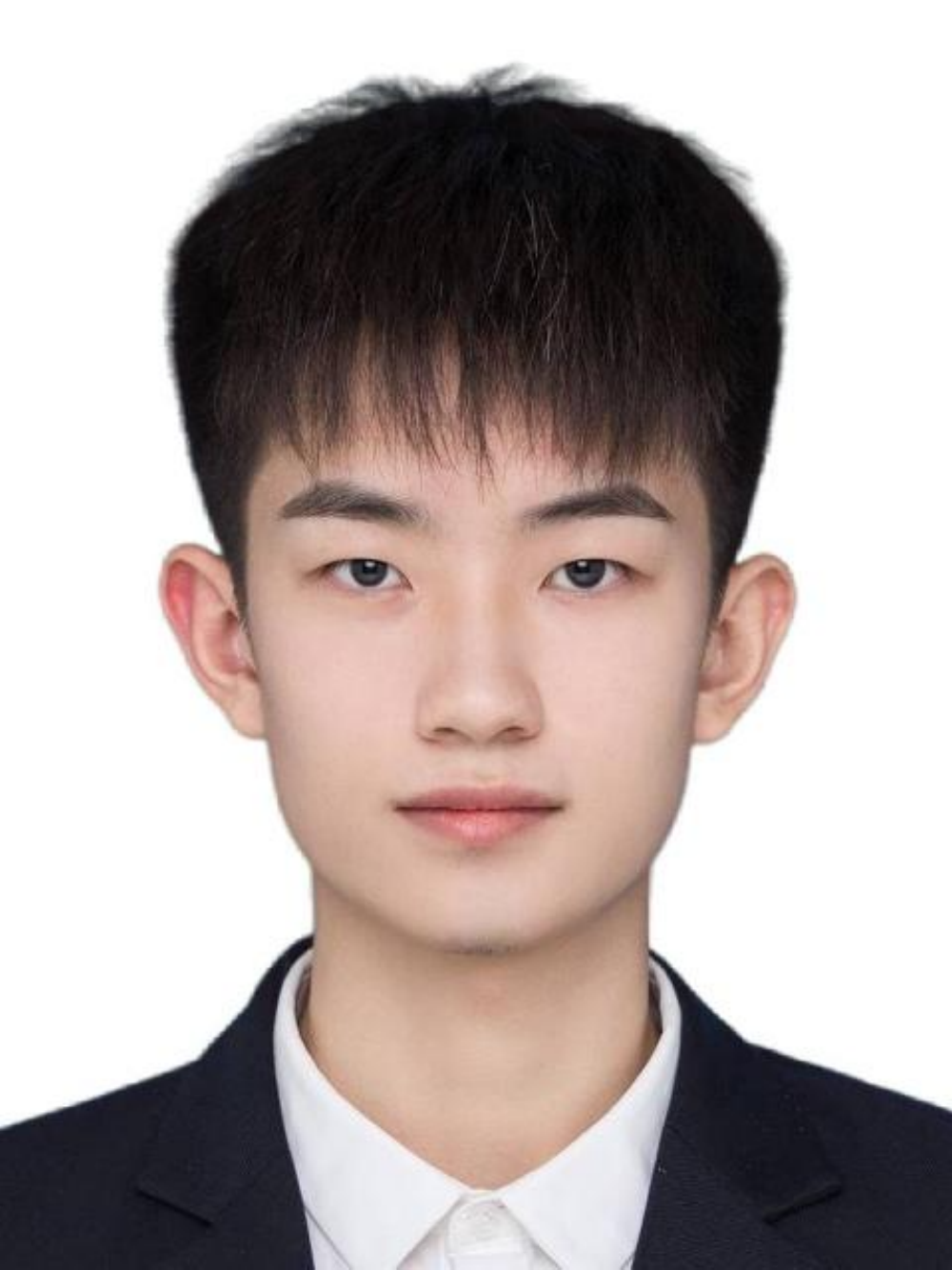}}]{Wenhan Chang} received his B.Eng. degree in 2022 and M.Eng. degree in 2025 from China University of Geosciences, Wuhan, China. He is currently pursuing a Ph.D. degree at School of Information Engineering, Zhongnan University of Economics and Law, China. His research interests include security and privacy preserving in deep learning.
\end{IEEEbiography}

\begin{IEEEbiography}
[{\includegraphics[width=1in,height=1.25in,clip,keepaspectratio]{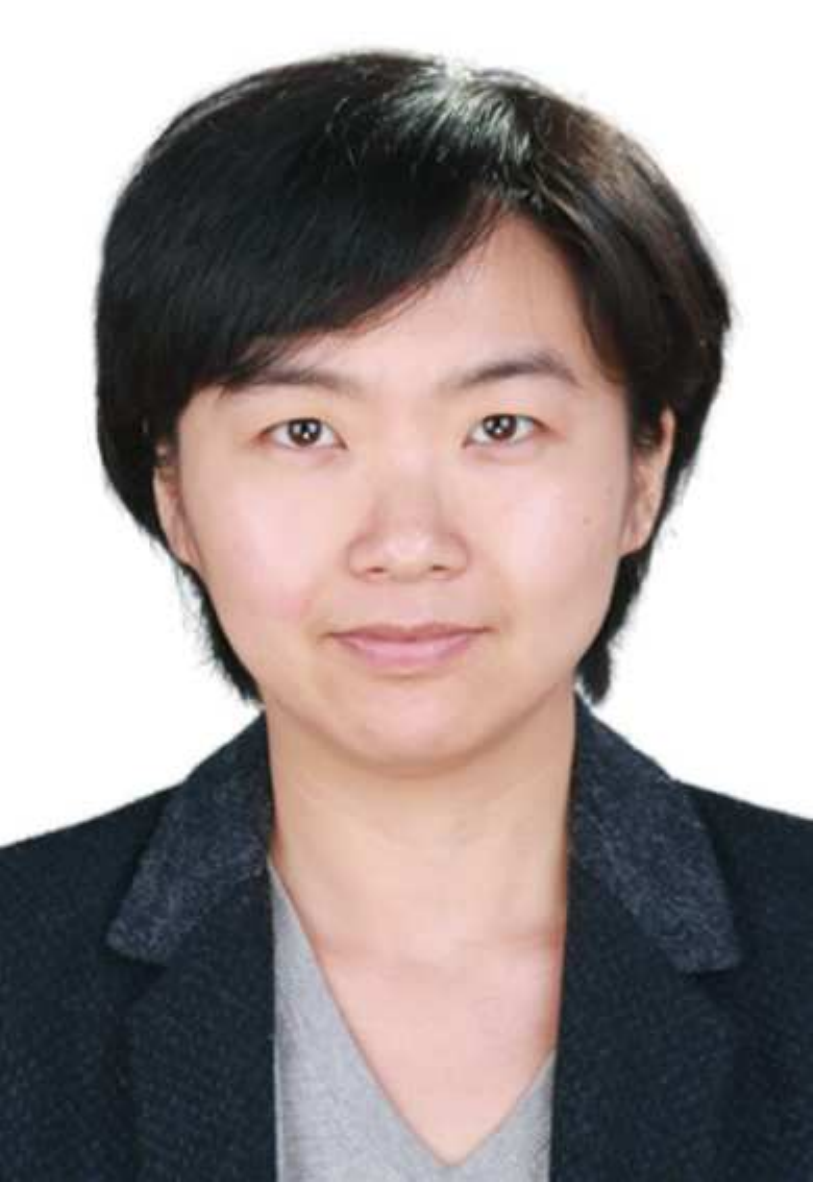}}]
{Tianqing Zhu} received her B.Eng. degree and her M.Eng. degree from Wuhan University, China, in 2000 and 2004, respectively. She also holds a PhD in computer science from Deakin University, Australia (2014). 
She is currently the Vice Dean of Institute of Data Science, City University of
Macau, Macao SAR, China. Prior to that, she was a lecturer with the School of Information Technology, Deakin University, and later an associate professor at University of Technology Sydney. Her research interests include privacy preserving, AI security and privacy, and network security.
\end{IEEEbiography}

\begin{IEEEbiography}
[{\includegraphics[width=1in,height=1.25in,clip,keepaspectratio]{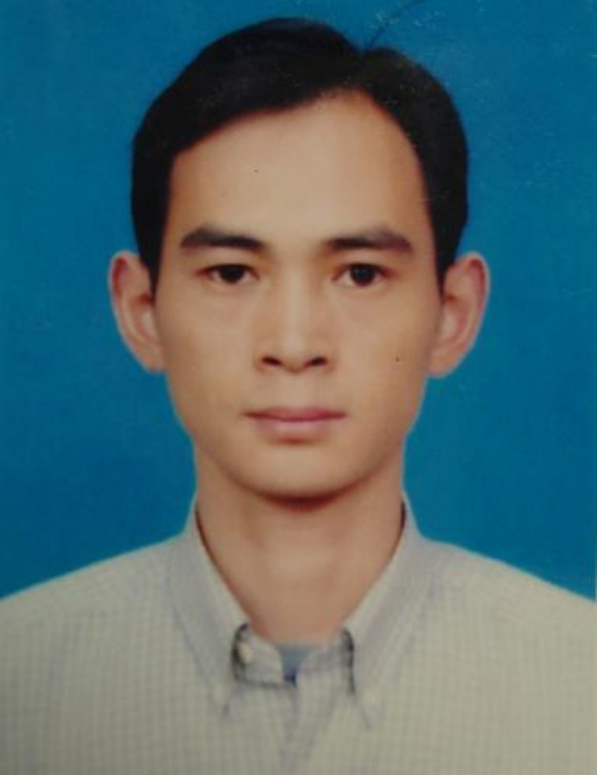}}]
{Ping Xiong} received his B.Eng degree from LanZhou Jiaotong University, China in 1997. He received his MEng and PhD degrees from Wuhan University, China, in 2002 and 2005, respectively. He is currently the professor of School of Information Engineering, Zhongnan University of Economics and Law, China. His research interests are information security, machine learning and privacy preservation.
\end{IEEEbiography}

\begin{IEEEbiography}
[{\includegraphics[width=1in,height=1.25in,clip,keepaspectratio]{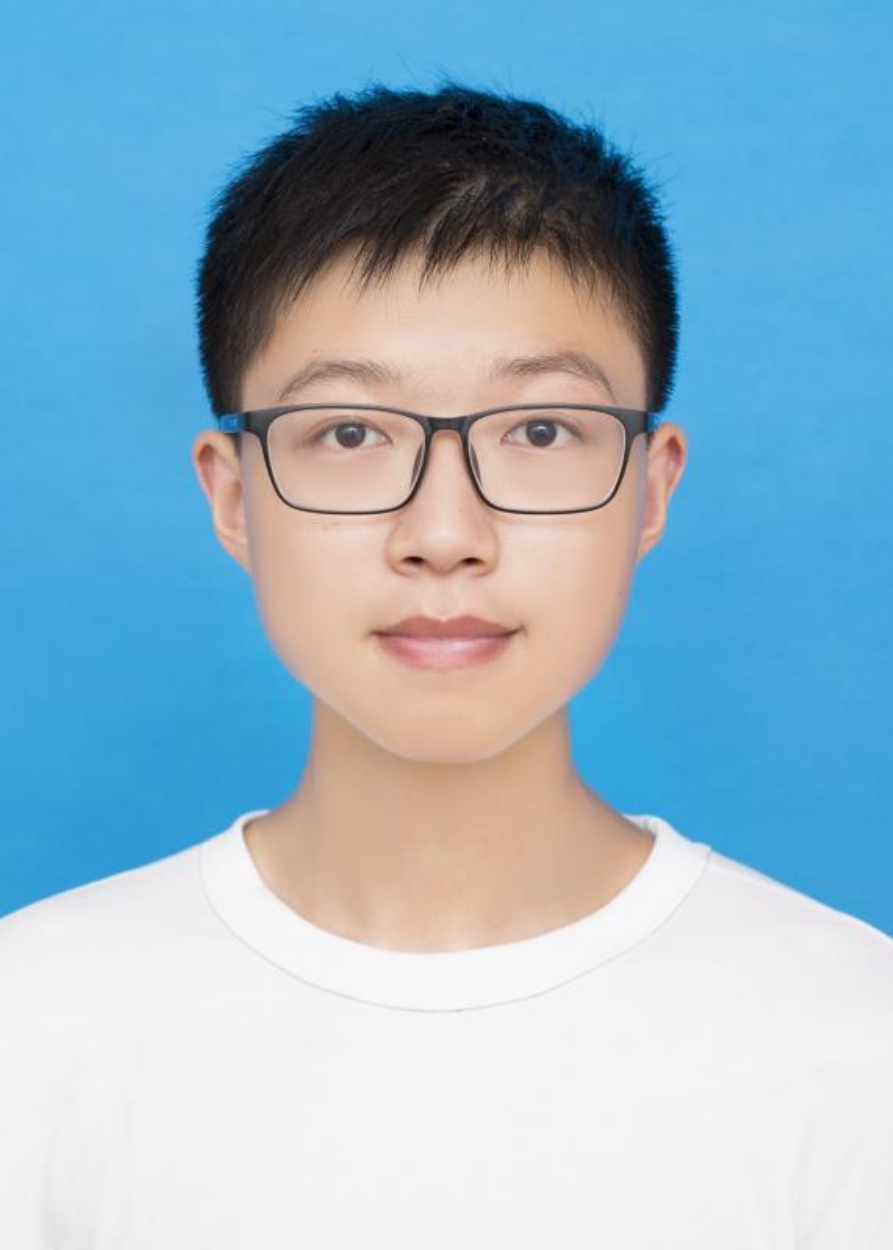}}]
{Yufeng Wu} received his B.Eng degree in Management from Wuhan University of Science and Technology in 2023. Currently, he is pursuing a Master's degree at China University of Geosciences, Wuhan, China. His research interests focus on model interpretability in image classification and privacy preservation.
\end{IEEEbiography}

\begin{IEEEbiography}
[{\includegraphics[width=1in,height=1.25in,clip,keepaspectratio]{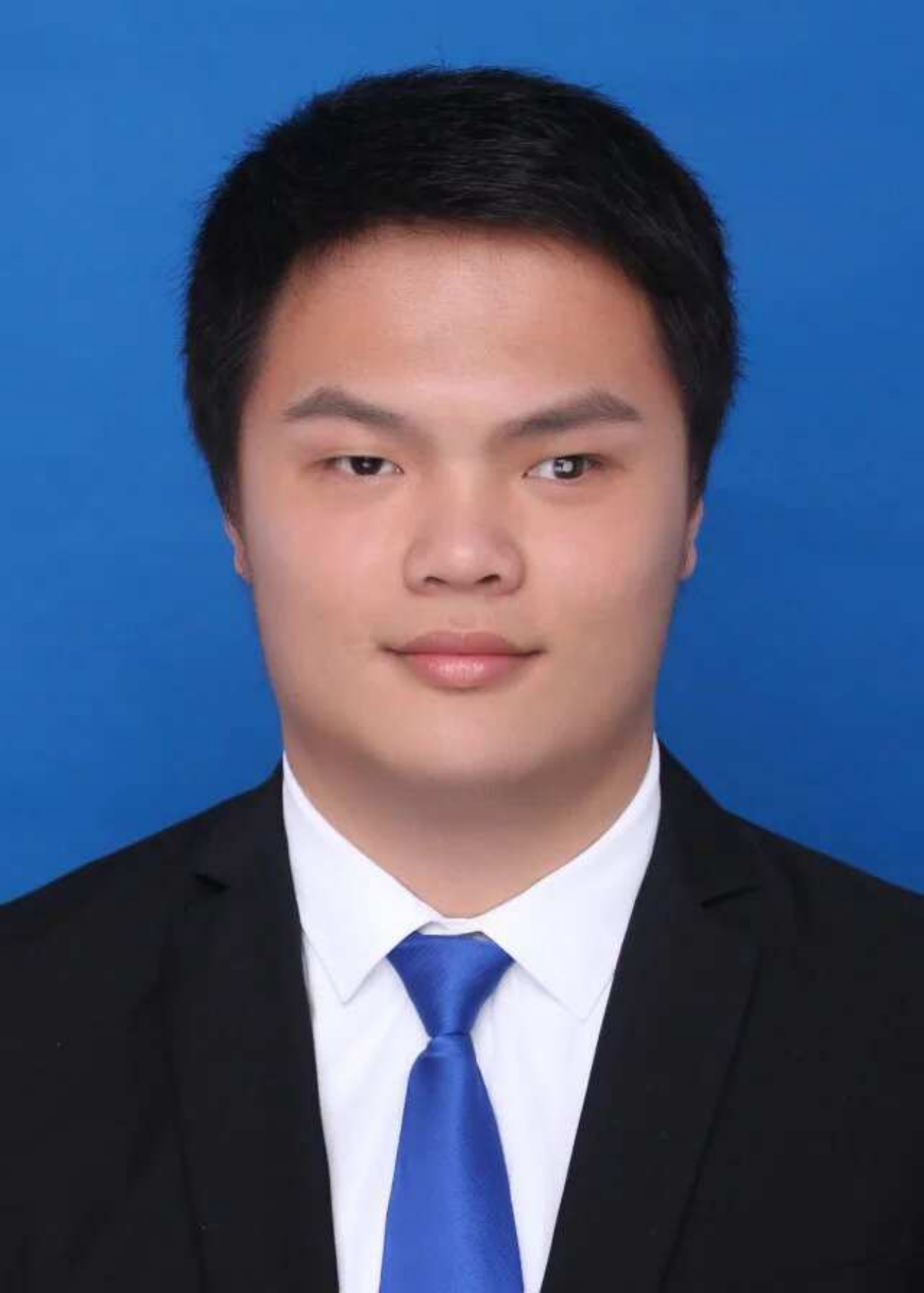}}]
{Faqian Guan} received the B.Eng. degree from Minnan Normal University, China, in 2017, and the M.Eng. degree from Fuzhou University, China, in 2020. He also holds a Ph.D. in Computer Science from China University of Geosciences, China. He is currently an Assistant Professor at the City University of Macau, Macao, China. His research interests include security and privacy in machine learning, and graph neural network. 
\end{IEEEbiography}

\begin{IEEEbiography}
[{\includegraphics[width=1.1in,height=1.25in,clip,keepaspectratio]{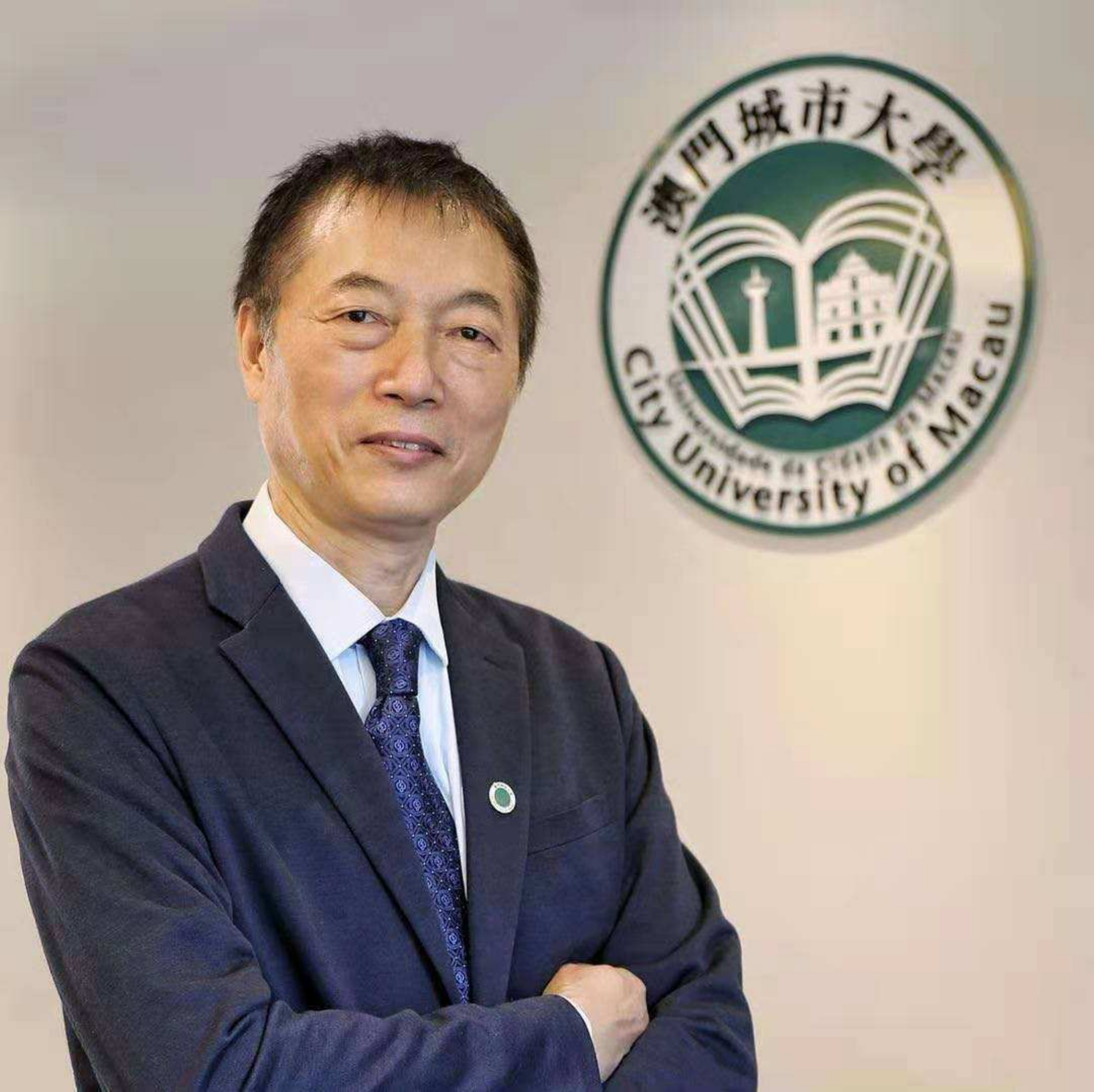}}]
{Wanlei Zhou} received the B.Eng and M.Eng degrees from Harbin Institute of Technology, Harbin, China in 1982 and 1984, respectively, and the PhD degree from The Australian National University, Canberra, Australia, in 1991, all in Computer Science and Engineering. He also received a DSc degree from Deakin University in 2002. He is currently the Vice Rector and Dean of Institute of Data Science, City University of Macau, Macao SAR, China. He has authored or coauthored more than 400 papers in refereed international journals and refereed international conferences proceedings, including many articles in IEEE transactions and journals. His research interests include security and privacy, parallel and distributed systems, and e-learning.
\end{IEEEbiography}

\vfill

\newpage

\maketitle

\end{document}